\documentclass[letterpaper]{article} 
\usepackage[draft]{aaai2026}  
\usepackage{times}  
\usepackage{helvet}  
\usepackage{courier}  
\usepackage[hyphens]{url}  
\usepackage{graphicx} 
\usepackage{natbib}  
\usepackage{caption} 
\usepackage{algorithm}
\usepackage{algorithmicx}
\usepackage{algpseudocode}
\usepackage{newfloat}
\usepackage{listings}
\DeclareCaptionStyle{ruled}{labelfont=normalfont,labelsep=colon,strut=off} 
\floatstyle{ruled}
\newfloat{listing}{tb}{lst}{}
\floatname{listing}{Listing}
\usepackage{amsmath} 
\usepackage{amssymb}
\usepackage{booktabs}
\usepackage{multirow}
\usepackage{array}
\usepackage[most]{tcolorbox}
\usepackage{subfig}
\usepackage{booktabs} 
\usepackage{amsmath} 
\usepackage{multirow} 
\usepackage{array} 
\usepackage{makecell} 

\author{
    Marco Valentino$^{1}$, Geonhee Kim$^{2}$, Dhairya Dalal$^{3}$, Zhixue Zhao$^{1}$, Andr\'e Freitas$^{2,4,5}$
}

\affiliations{
$^{1}$University of Sheffield, UK \quad
 $^{2}$University of Manchester, UK \quad
    $^{3}$University of Galway, Ireland \\
    $^{4}$Idiap Research Institute, Switzerland \quad
    $^{5}$National Biomarker Centre, CRUK-MI, UK\\
    m.valentino@sheffield.ac.uk
}

\usepackage{bibentry}

\usepackage{booktabs} 
\usepackage{amsmath} 
\usepackage{multirow} 
\usepackage{makecell} 

\newcommand{\ours}[0]{\texttt{K-CAST}}
\newcommand{\cast}[0]{\texttt{CAST}}

\title{Mitigating Content Effects on Reasoning in Language Models Through Fine-Grained Activation Steering}

\begin{document}

\maketitle

\begin{abstract}
Large language models (LLMs) exhibit reasoning biases, often conflating content plausibility with formal logical validity. This can lead to wrong inferences in critical domains, where plausible arguments are incorrectly deemed logically valid or vice versa. 
This paper investigates how content biases on reasoning can be mitigated through activation steering, an inference-time technique that modulates internal activations. 
Specifically, after localising the layers responsible for formal and plausible inference, we investigate activation steering on a controlled syllogistic reasoning task, designed to disentangle formal validity from content plausibility. 
An extensive empirical analysis reveals that contrastive steering methods consistently support linear control over content biases. However, a static approach is insufficient to debias all the tested models. We then investigate how to control content effects by dynamically determining the steering parameters through fine-grained conditional methods. By introducing a novel kNN-based conditional approach (\ours{}), we demonstrate that conditional steering can effectively reduce biases on unresponsive models, achieving up to 15\% absolute improvement in formal reasoning accuracy. Finally, we found that steering for content effects is robust to prompt variations, incurs minimal side effects on multilingual language modeling capabilities, and can partially generalize to different reasoning tasks. In practice, we demonstrate that activation-level interventions offer a scalable inference-time strategy for enhancing the robustness of LLMs, contributing towards more systematic and unbiased reasoning capabilities.

\end{abstract}

\begin{links}
 \link{Code and data}{https://github.com/neuro-symbolic-ai/steering_content_effects}
\end{links}

\section{Introduction}

Large language models (LLMs) possess advanced natural and common-sense reasoning capabilities but are prone to content effects -- i.e., systematic biases where prior knowledge and believability of content influence logical inference~\citep{bertolazzi-etal-2024-systematic,10.1093/pnasnexus/pgae233}. For example, an LLM may incorrectly judge a logically invalid syllogism as valid if its content aligns with common-sense knowledge (e.g., “All students read; some readers are professors; therefore some students are professors”), mirroring human content biases~\cite{10.1093/pnasnexus/pgae233}. Such behaviour violates the requirement of formal reasoning, where validity should depend only on logical form, not content. Recent studies have documented these content-based reasoning failures~\cite{bertolazzi-etal-2024-systematic}, showing that models find factually believable premises easier to “prove” and struggle with abstract or counter-intuitive ones~\cite{10.1093/pnasnexus/pgae233}. This undermines LLMs' reliability on formal reasoning, particularly logic-oriented tasks highlighted by \citet{bertolazzi-etal-2024-systematic}.

On the other hand, prompting strategies alone are insufficient to eliminate content effects. 
Chain-of-thought (CoT) prompting and related methods \cite{wei2022chain,kojima2022large} can improve reasoning. However, biases often persist in the generated explanations, and models may still arrive at content-biased conclusions even when “thinking aloud”~\cite{ranaldi2025improving}. In fact, \cite{bertolazzi-etal-2024-systematic} finds that while CoT and fine-tuning boost accuracy on logical deductions, they do not fully remove biases like content believability effects.  Similarly, neuro-symbolic approaches have been proposed to improve robustness in formal reasoning with LLMs \cite{quan-etal-2024-verification,quan-etal-2025-faithful,quan2025peirce,pan-etal-2023-logic,lyu-etal-2023-faithful}. However, they introduce the complexity of integrating LLMs with external symbolic solvers.

\begin{figure*}[t]
\centering
\includegraphics[width=\textwidth]{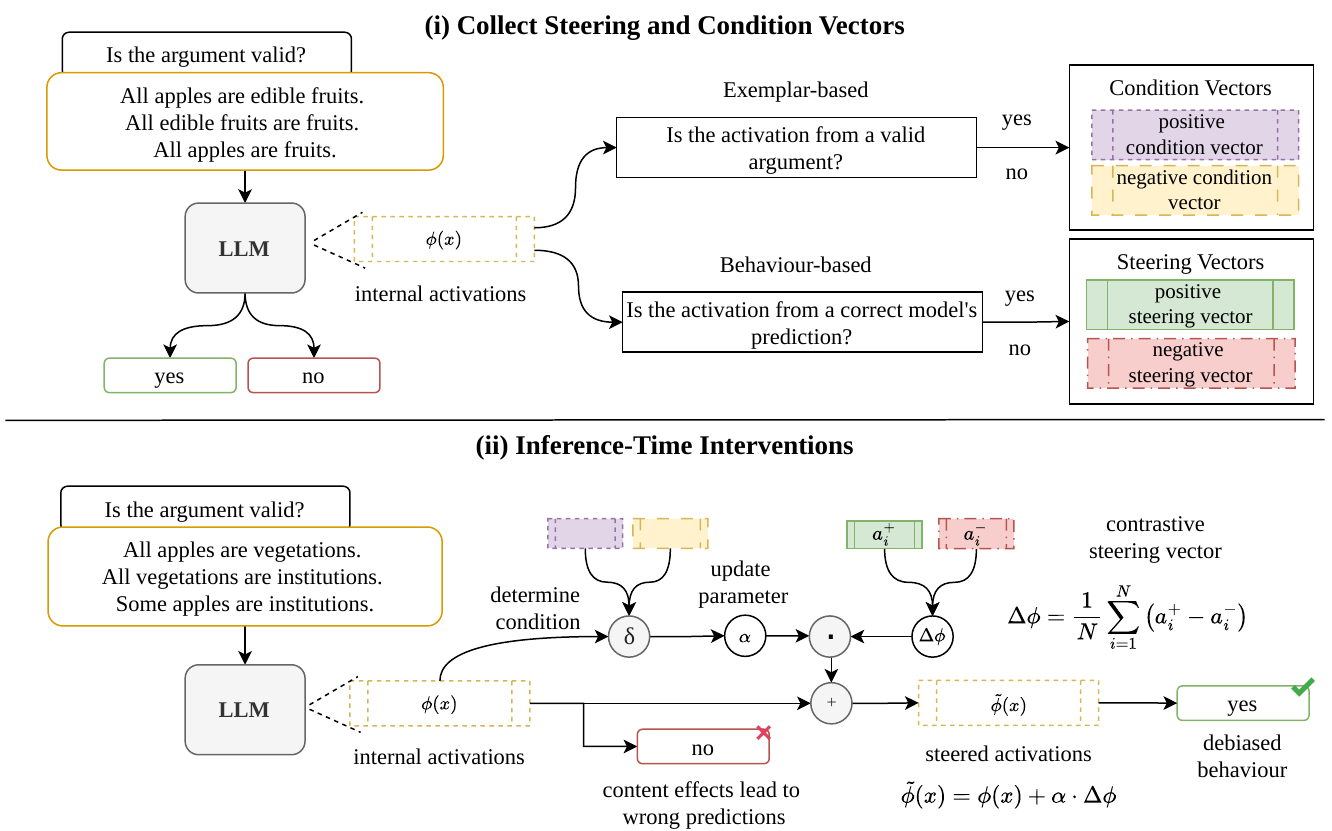}
\caption{Overview of our methodology for mitigating content effects on reasoning via activation steering. We first curate a controlled syllogistic reasoning dataset designed to disentangle formal validity from content plausibility. Subsequently, after localising the layers mostly responsible for formal and plausible inference through probing, we investigate static and conditional steering methods to debias models' behaviour.}
\label{fig:pipeline}
\end{figure*}

Unlike existing methods, we directly manipulate internal activations to investigate if content-invariance can be effectively enforced through test-time interventions and to gain a deeper understanding of the internal representational mechanisms (see Figure \ref{fig:pipeline}). 
Overall, our contribution and findings can be summarised as follows:


\paragraph{A large-scale dataset to disentangle content from formal reasoning.} Expanding on previous work \cite{bertolazzi-etal-2024-systematic,kim2024mechanistic,wysocka-etal-2025-syllobio}, we generate a synthetic dataset leveraging known syllogistic arguments, considering the intersection of plausible/implausible and formally valid/invalid arguments. The dataset includes over 16k arguments generated by instantiating abstract syllogistic schemes with the support of Wordnet \cite{miller1995wordnet}.

\paragraph{Localizing formal and plausible inference.} We perform an observational study through probing \cite{ferreira2021does,belinkov2022probing} to localise information about the validity and plausibility of arguments within the models. The experiments reveal that the information is maximally localised in later layers, peaking at the third quarter of the layers in the residual stream across different LLMs. 

\paragraph{Evaluating static contrastive steering methods.} Leveraging the observational study, we investigate static and contrastive activation steering methods \cite{panickssery2023steering}. In general, we found that contrastive steering is effective on most of the tested models. In particular, the experiments reveal that steering vectors can explicitly control models' output along a linear direction depending on the steering parameters, influencing the accuracy on both valid and invalid arguments. However, we found that static steering cannot improve performance on all the tested models.

\paragraph{Adapting and introducing fine-grained conditional steering methods.}  We adapt the recently proposed conditional activation steering (\cast{}) method \cite{mallen2024cast} for content effects, and propose a new fine-grained variation employing a k-NN classifier to dynamically determine the steering parameters (\ours{}). We found that such methods can reduce biases on models that are unresponsive to static steering while, at the same time, increasing overall accuracy by up to $\approx$ 15\% absolute value.

\paragraph{Robustness analysis and out-of-distribution generalisation.} We investigate the impact of steering for content effects on multilingual language modeling capabilities \cite{raffel2020exploring} and out-of-distribution reasoning tasks \cite{PrOntoQA,chan2024rulebreakers}. We found that steering is well-localized, incurring minimal side effects on language modeling capabilities. At the same time, we found that steering vectors computed on the synthetic data can generalize to some extent to different reasoning tasks, with some variations across models. These results highlight both the potential of steering to improve targeted reasoning capabilities as well as the persisting challenges in enabling full generalization.

\section{Background}

Recent research has demonstrated that LLMs exhibit \emph{content effects} in formal reasoning tasks, mirroring cognitive biases that may align or differ from those observed in humans \cite{10.1093/pnasnexus/pgae233,kim2024mechanistic,mondorf2024comparing,seals2024evaluating,wysocka-etal-2025-syllobio,eisape2024systematic}. These effects arise when the semantic plausibility of a prompt influences the model’s reasoning process, often leading to correct conclusions for plausible statements and systematic errors for implausible but logically valid ones \cite{bertolazzi-etal-2024-systematic}.

\cite{10.1093/pnasnexus/pgae233} first showed that LLMs perform better on reasoning tasks when the content of the problem aligns with world knowledge. In their experiments, models were significantly more accurate on syllogistic tasks when the conclusions were semantically plausible, even when this plausibility conflicted with the actual logical validity of the argument. This indicates a bias toward material reasoning (reasoning grounded in semantic associations) rather than formal reasoning (reasoning based strictly on logic).

Further work by~\cite{bertolazzi-etal-2024-systematic} systematically evaluated LLMs on a broad suite of syllogisms and found that performance dropped sharply for arguments that contradicted commonsense knowledge. This reliance on content plausibility suggests that LLMs are susceptible to semantic interference, failing to uphold the norms of formal logic when they conflict with prior knowledge.
\cite{seals2024evaluating} also emphasized this discrepancy, showing biases in LLMs' deductive reasoning. To the best of our knowledge, this is the first work investigating how content effects can be reduced through activation steering techniques \cite{subramani2022extracting,hernandezinspecting,zou2023representation,turner2023steering,li2023inference,zhao2024steering}.

\section{Methodology}

Our goal is to investigate and mitigate \emph{content effects} in LLMs, i.e., systematic biases where semantic plausibility influences logical reasoning.
To this end, we design a controlled syllogistic reasoning task, leveraging abstract syllogistic schemes automatically instantiated through taxonomic knowledge from external knowledge bases (Sec.~\ref{sec:method_eval}) and apply activation-level steering techniques to modulate model behavior toward formal validity assessment (Sec.~\ref{sec:method_steering}). Further, we identify the limitations of current state-of-the-art steering methods, and propose a more fine-grained steering approach (Sec.~\ref{sec:method_steering_knn_cast}). Figure \ref{fig:pipeline} provides a high-level overview of the methodology.

\subsection{
Formal Syllogistic Reasoning}
\label{sec:method_eval}

Inspired by recent work~\cite{bertolazzi-etal-2024-systematic,wysocka-etal-2025-syllobio,kim2024mechanistic,10.1093/pnasnexus/pgae233}, we evaluate formal reasoning in LLMs through syllogistic arguments. We formalise the task of syllogistic reasoning as a binary classification problem, where the objective is to determine the \emph{formal validity} of a syllogism, whether its conclusion follows logically from its premises, irrespective of the plausibility of the content. Specifically, the model is expected to predict \texttt{VALID} or \texttt{INVALID} based solely on the logical form in the syllogism structure $\mathcal{S}$. 

A syllogism $\mathcal{S}$ is defined as a triple $\mathcal{S} = (P_1, P_2, C)$
where $P_1$ and $P_2$ are the two categorical premises, and $C$ is the conclusion. Each statement is expressed in natural language and conforms to standard syllogistic forms (e.g., universal affirmative, universal negative, particular affirmative).

\paragraph{Controlling plausibility for content effect evaluation.} To isolate formal validity from world knowledge, we design the task to include the following types of syllogistic arguments:
\begin{itemize}
    \item \textbf{Plausible Valid:} All apples are edible fruits. All edible fruits are fruits. All apples are fruits.
    \item  \textbf{Implausible Valid:} All apples are vegetations. All vegetations are institutions. Some apples are institutions.
    \item \textbf{Plausible Invalid:} All apples are edible fruits. All apples are fruits. All edible fruits are fruits.
    \item \textbf{Implausible Invalid:} All apples are institutions. All vegetations are institutions. All apples are vegetations.
\end{itemize}

This setup allows us to decouple reasoning based on logical form from reasoning based on material content. Models demonstrating robust formal reasoning will maintain consistent accuracy across plausible and implausible conditions by focusing exclusively on argument structure.

\paragraph{Syllogistic arguments generation.}
We construct a dataset of approximately 16,000 syllogistic arguments in English to systematically analyze content effects. Each argument instantiates one of the formal syllogistic schemas (details in the supplementary material) and is explicitly varied along dimensions of \emph{formal validity} and \emph{semantic plausibility}.

The data generation process begins with the formalization of syllogistic structures in first-order logic (FOL), following prior work on logical reasoning datasets \citep{bertolazzi-etal-2024-systematic, wysocka-etal-2025-syllobio}.
Each logical schema is then converted into natural language templates such as: 
\emph{All A are B, All B are C, All A are C}.

To control semantic content, we instantiate these syllogistic templates with concrete noun phrases drawn from WordNet\footnote{\url{https://wordnet.princeton.edu/}}\cite{miller1995wordnet} based on taxonomic hierarchies, using hypernym-hyponym relations between concepts.

\begin{figure*}[t!]
\centering
{\includegraphics[width=\textwidth]{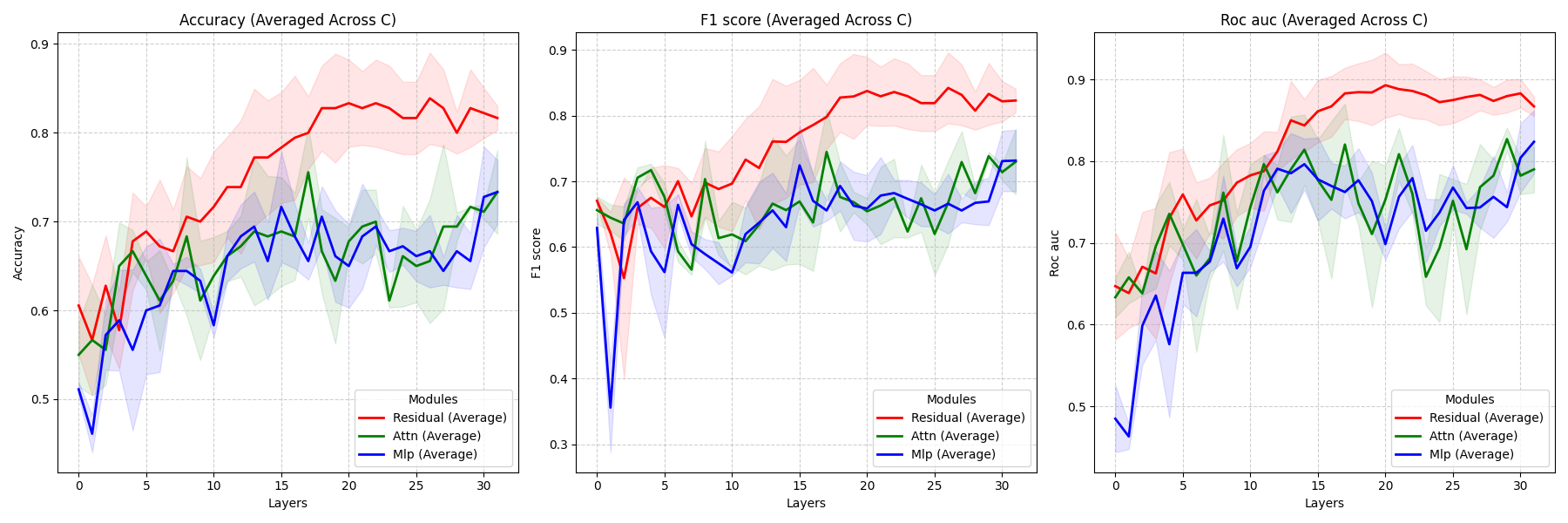}}
\caption{Linear probing results for formal validity on Llama-3.1 8b. The probing experiments reveal that the information about validity is maximally localised in the residual stream of later layers across different LLMs.}
\label{fig:probing}
\end{figure*}

\begin{table*}[t]
\centering
\small
\begin{tabular}{ll|ccc|ccc|c|c|c}
\toprule
\multirow{2}{*}{\textbf{Model}} & \multirow{2}{*}{\textbf{Size}} & \multicolumn{3}{c|}{\textbf{Base Model ($\alpha=0$)}} & \multicolumn{3}{c|}{\textbf{Steered Model ($\alpha_{best}$)}} & \multirow{2}{*}{$\mathbf{\alpha_{\text{best}}}$} & \multirow{2}{*}{\thead{\textbf{ $\Delta_{abs}^{Acc/CE}$}}} & \multirow{2}{*}{\thead{\textbf{ $\Delta_{rel}^{Acc/CE}$}}} \\
\cline{3-8}
& & \textbf{Acc $\uparrow$} & \textbf{CE $\downarrow$} & \textbf{Acc/CE $\uparrow$} & \textbf{Acc $\uparrow$} & \textbf{CE $\downarrow$} & \textbf{Acc/CE $\uparrow$} & & & \rule{0pt}{1.3em}\\ 
\midrule
\multicolumn{11}{l}{\textbf{Zero-shot}} \\
\midrule
\textbf{Llama 3.2} & 1b & 58.17 & 44.04 & 1.32 & 73.56 & \textbf{6.35} & \textbf{11.58} & -0.9 & $\mathbf{10.26}$ & $\mathbf{777.27}$ \\
& 3b & 77.79 & \textbf{17.50} & \textbf{4.45} & 77.79 & 17.50 & 4.45 & 0.0 & $0.00$ & $0.00$ \\
\textbf{Llama 3.1} & 8b & \textbf{78.27} & 30.77 & 2.54 & \textbf{85.10} & 14.04 & 6.06 & 0.9 & $3.52$ & $138.58$ \\
\midrule
\textbf{Gemma 2} & 2b & 73.27 & 32.43 & 2.26 & 74.13 & 20.83 & 3.56 & 1.8 & $1.30$ & $57.52$ \\
& 9b & \textbf{85.00} & \textbf{8.46} & \textbf{10.05} & \textbf{83.27} & \textbf{1.92} & \textbf{43.37} & 0.6 & $\mathbf{33.32}$ & $\mathbf{331.54}$ \\
\midrule
\textbf{Qwen 2.5} & 1.5b & 75.67 & 14.42 & 5.25 & 77.79 & 12.88 & 6.04 & 0.3 & $0.79$ & $15.05$ \\
& 3b & 85.29 & 7.12 & \textbf{11.99} & 85.29 & 7.12 & 11.99 & 0.0 & $0.00$ & $0.00$ \\
& 7b & \textbf{88.85} & \textbf{5.39} & 16.48 & \textbf{89.90} & \textbf{0.96} & \textbf{93.65} & -1.5 & $\mathbf{77.17}$ & $\mathbf{468.26}$ \\
\midrule
\multicolumn{11}{l}{\textbf{Few-shot}} \\
\midrule
\textbf{Llama 3.2} & 1b & 57.21 & 45.70 & 1.25 & 66.44 & \textbf{6.08} & \textbf{4.94} & -1.5 & $\mathbf{3.69}$ & $\mathbf{295.20}$ \\
& 3b & \textbf{72.79} & 28.14 & \textbf{2.59} & \textbf{78.27} & 22.31 & 3.51 & 0.3 & $0.92$ & $35.52$ \\
\textbf{Llama 3.1} & 8b & 40.58 & \textbf{20.26} & 2.00 & 30.67 & 14.81 & 2.07 & 0.3 & $0.07$ & $3.50$ \\
\midrule
\textbf{Gemma 2} & 2b & 69.42 & \textbf{14.23} & 4.88 & 70.00 & 12.31 & 5.69 & -0.3 & $0.81$ & $16.60$ \\
& 9b & \textbf{84.61} & 15.25 & \textbf{5.54} & \textbf{80.38} & \textbf{3.33} & \textbf{24.14} & 0.3 & $\mathbf{18.60}$ & $\mathbf{335.74}$ \\
\midrule
\textbf{Qwen 2.5} & 1.5b & 51.92 & 65.12 & 0.80 & 72.69 & 32.18 & 2.26 & -1.2 & $1.46$ & $182.50$ \\
& 3b & 86.44 & 13.91 & 6.21 & 86.63 & \textbf{4.80} & 18.02 & 0.6 & $\mathbf{11.81}$ & $\mathbf{190.18}$ \\
& 7b & \textbf{89.80} & \textbf{9.36} & \textbf{9.60} & \textbf{89.90} & 4.81 & \textbf{18.70} & 0.9 & $9.10$ & $94.79$ \\
\bottomrule
\end{tabular}
\caption{Results of contrastive steering with static values of $\alpha$. The table compares the performance of unsteered base models ($\alpha=0$) against the optimal steered performance ($\alpha_{\text{best}}$) with values of $\alpha$ selected from the interval  $[-3.0, 3.0]$. $\text{Acc/CE}$ is the composite metric (Accuracy/Content Effect). The final columns quantify the $\text{Acc/CE}$ gain: $\Delta_{\text{abs}}$ is the raw increase, and $\Delta_{\text{rel}}$ is the percentage increase relative to the baseline. The results reveal that contrastive steering is highly effective for most models, except Llama 3.2 3b and Qwen 2.5 3b.}
\label{tab:res_static_steering}
\end{table*}

\subsection{Activation Steering}\label{sec:method_steering}

Activation steering is a causal intervention technique for modulating the internal computation of LLMs by linearly modifying hidden activations, also known as activation engineering~\cite{subramani2022extracting,hernandezinspecting,zou2023representation,turner2023steering,li2023inference,zhao2024steering}. In this work, we adopt both static and conditional activation steering methods.

\paragraph{Contrastive Activation Addition (CAA)} computes the steering vector using a set of labelled examples, based on the observed model behavior ~\cite{panickssery2023steering}. 

Let $\phi(x) \in \mathbb{R}^d$ denote the activation vector at a chosen layer and token position for input $x$. Given a set of $N$ contrastive pairs of activations $\mathcal{P} = \{(a_i^+, a_i^-)\}_{i=1}^N$, where $a_i^+ = \phi(x_i^+)$ and $a_i^- = \phi(x_i^-)$ are positive and negative activation vectors derived from inputs $x_i^+$ and $x_i^-$ leading to desired and contrasting behaviors respectively.  The resulting steering vector is the mean difference between positive and negative activations:
\begin{equation}
\small
\Delta \phi = \frac{1}{N} \sum_{i=1}^{N} \left( a_i^+ - a_i^- \right)
\end{equation}

At inference time, given a new input $x$, the model is steered by modifying its internal activations $\tilde{\phi}(x) = \phi(x) + \alpha \cdot \Delta \phi$, where $\alpha$ is a scaling hyperparameter determining the strength of the intervention.

We apply CAA to reduce \emph{content effects} by steering activations toward representations associated with content-invariant outputs. To this end, the positive vectors are collected from inputs leading to correct formal validity predictions, while the negative vectors are collected from incorrect predictions affected by content bias.

\paragraph{Conditional Activation Steering (CAST)} is a steering method designed to enable selective modulation of model behaviors by conditionally applying activation steering based on the input context~\cite{mallen2024cast}. Unlike traditional activation steering methods that uniformly apply steering vectors across all inputs, CAST introduces a mechanism to determine, at inference time, whether to apply a steering vector based on the similarity of the current input's activation to predefined condition vectors.

Formally, let $\phi(x) \in \mathbb{R}^d$ denote the activation vector at a specified layer and position for input $x$. Given a labeled dataset $\mathcal{D} = \{(x_i, y_i)\}_{i=1}^N$, where $y_i \in \{+1, -1\}$ denotes the presence or absence of a target condition on  $x_i$ (i.e., the argument is formally valid), \cast{} computes a condition vector $\psi_{c}$ based on the average aggregation (or PCA) of individual activations vectors $\phi(x_i)$ such that $y_i = +1$. During inference, for a given input $x$, the similarity between $\phi(x)$ and $\pi_{\psi_c}(\phi(x))$ -- i.e., the projection of $\phi(x)$ onto $\psi_c$ -- is computed:

\begin{equation}
\small
\text{sim}(\phi(x), \pi_{\psi_c}(\phi(x))) = \frac{\phi(x) \cdot \pi_{\psi_c}(\phi(x))}{\|\phi(x)\| \|\pi_{\psi_c}(\phi(x))\|}
\end{equation}

In the standard \cast{} method, if the similarity exceeds a predefined threshold $\theta_c$, a corresponding steering vector $\Delta \phi_c$ is applied $
\tilde{\phi}(x) = \phi(x) + \alpha \cdot \Delta \phi_c$,
where $\alpha$ is a scaling parameter controlling the strength of the intervention.

In this work, we adapt \cast{} to dynamically determine the value of the scaling parameter $\alpha$, since our empirical analysis on static steering reveals that the sign of $\alpha$ enables explicit control over the accuracy on valid and invalid arguments. In particular, given two condition vectors $\psi_{c_+}$ and $\psi_{c_-}$, the first computed for \emph{valid} arguments and the second computed for \emph{invalid} arguments, we modify the value of $\alpha$ dynamically according to the following function:

\begin{equation}
\small
\begin{split}
f(\alpha, \phi(x),\psi_{c_+}, \psi_{c_-}) = \\
\begin{cases}
-\alpha & \text{if } \text{sim}(\phi(x), \pi(\phi(x)_{\psi_{c_+}}))>\text{sim}(\phi(x), \pi(\phi(x)_{\psi_{c_-}})) \\
\alpha & \text{otherwise}
\end{cases}
\end{split}
\end{equation}

Therefore, we perform conditional steering via
$\tilde{\phi}(x) = \phi(x) + f(\alpha, \phi(x),\psi_{c_+}, \psi_{c_-}) \cdot \Delta \phi$, where $\Delta \phi$ is a standard contrastive steering vector.

\subsubsection{\ours{}: kNN-Based Conditional Activation Steering}\label{sec:method_steering_knn_cast}

One limitation of \cast{} is that the condition vectors $\psi_c$ are typically computed via aggregating individual activations from different training examples. This can cause a loss of information that undermines the ability to effectively determine the correct condition for test-time intervention. To address this, we introduce an extension to \cast{} that employs a k-Nearest Neighbors (kNN) approach for condition determination, thereby mitigating potential information loss from coarse-grained aggregation methods.

Given a labeled dataset $\mathcal{D} = \{(x_i, y_i)\}_{i=1}^N$, where $y_i \in \{+1, -1\}$ denotes the presence or absence of a target condition on  $x_i$ (i.e., the argument is formally valid), we proceed as follows:

\begin{enumerate}
    \item For each input $x_i$ in $\mathcal{D}$, compute and store an individual condition activation vector $\psi(x_i)_{y_i}$.
    \item At inference time, for a new input $x$, compute its activation vector $\phi(x)$.
    \item  Identify the set $\mathcal{N}_k(x) \subset \mathcal{D}$ of $k$ nearest neighbors to $\phi(x)$ based on cosine similarity.
    \item Determine the majority condition label $\hat{y}(x)$ among the neighbors: 
\begin{equation}
\small
\hat{y}(x) = \text{sign}\left( \sum_{(x_j, y_j) \in \mathcal{N}_k(x)} y_j \right)
\end{equation}
\small
    \item  Dynamically determining the steering parameters based on the majority condition label:
\begin{equation}
    \tilde{\phi}(x) = \phi(x) + f( \alpha, \Delta \phi, \hat{y}(x))
\end{equation}
\end{enumerate}

where $\Delta \phi$ is a standard contrastive steering vector. While this method can be used to arbitrarily adapt the steering method, similarly to \cast{}, we employ it to dynamically determine the value of $\alpha$ at test time via $\tilde{\phi}(x) = \phi(x) - \hat{y}(x) \cdot\alpha \cdot \Delta \phi_c$.

Compared to \cast, \ours{} allows for a more granular determination of how to apply the steering interventions, leveraging the local structure of the activation space in the training set.

\begin{figure*}[t]
\centering
\subfloat[Llama 3b (Static) \label{fig:acc_ce_llama_3b}]{\includegraphics[width=\columnwidth]
{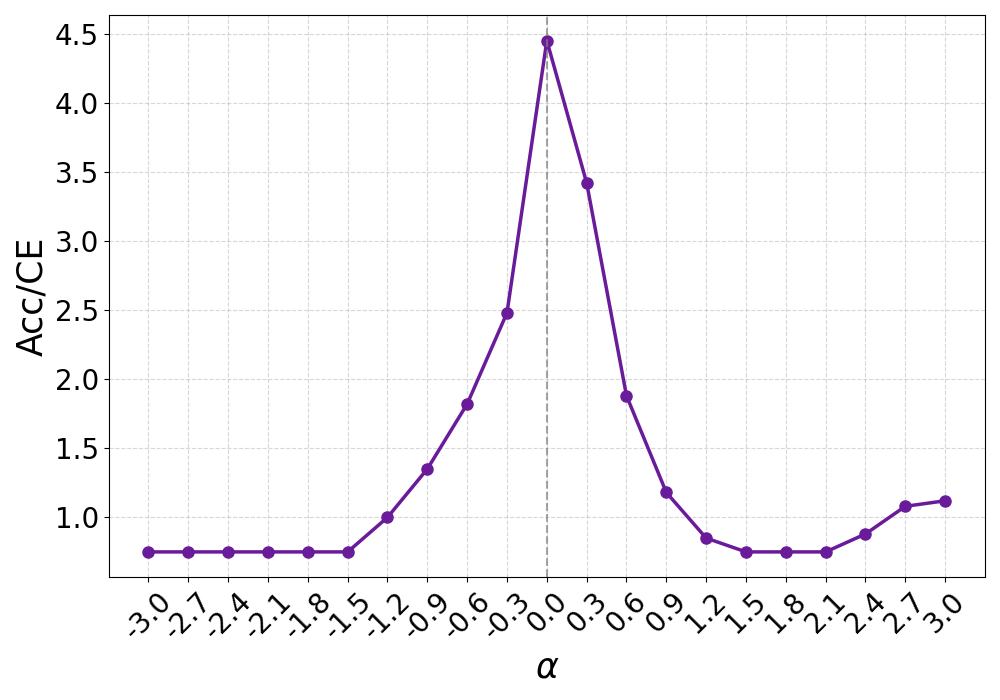}}
\subfloat[Llama 3b (\ours) \label{fig:acc_ce_knn}]{\includegraphics[width=\columnwidth]{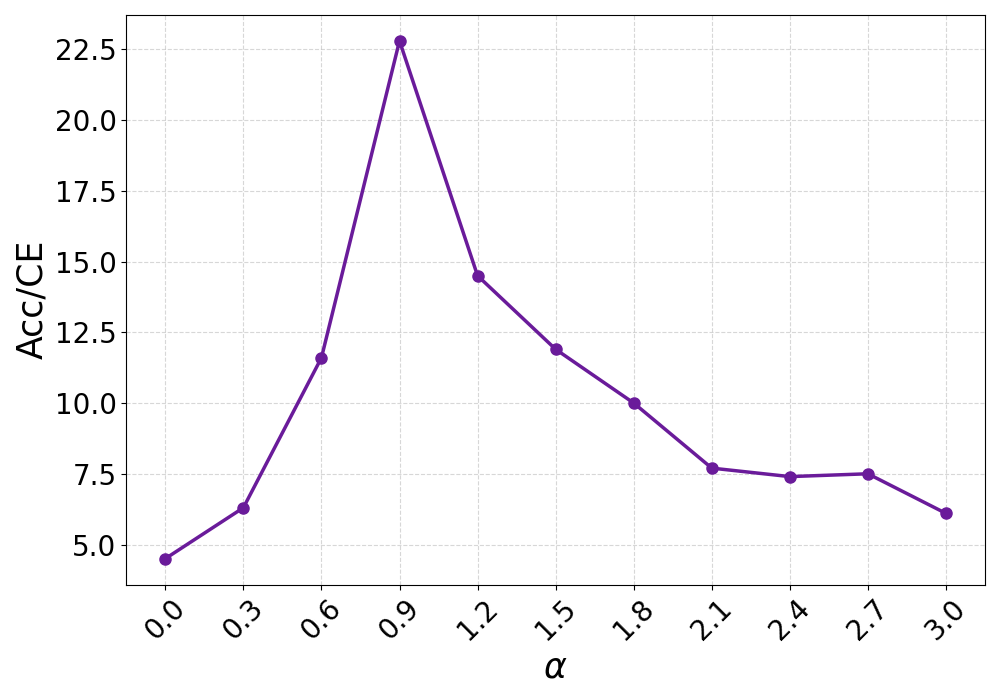}}
\caption{(a) Results of steering with static values of $\alpha$ on Llama 3b (note that $\alpha = 0$ represents the performance of the base model without steering). (b) Impact of conditional steering (\ours{}) on Llama 3b. \ours{} increases accuracy while reducing content effects on models that are unresponsive to a static steering approach.}
\label{fig:detailed_contrastive_steering_main}
\end{figure*}

\section{Empirical Evaluation}
\label{sec:results}

\paragraph{Models.}  We evaluate the steering performance on three model families, covering different spans of model sizes: Llama (3.2-1b-it, 3.2-3b-it, 3.1-8b) \cite{grattafiori2024llama}, Gemma-2 (2b-it, Gemma-2-9b-it) \cite{team2024gemma}, Qwen 2.5 (1.5b-it, 3b-it, 7b-it) \cite{bai2023qwen}. We use the instruction-tuned version of each model and evaluate both performance in zero-shot setting and in-context learning (ICL) via few-shot prompts, providing a total of 4 random examples from the training set.

\paragraph{Probing for content effects.} To inform subsequent test-time interventions and steering, we performed a preliminary observational study through linear probing \cite{belinkov2022probing,ferreira2021does} to identify where information about the validity and plausibility of arguments might be encoded within the models. To this end, we employ a linear layer on top of the models' frozen activations after processing a syllogistic argument, classifying whether the argument is valid/invalid and plausible/implausible. Overall, the probing experiments reveal that the information for validity and plausibility is maximally localised in later layers in the residual stream, consistently peaking at the third quarter of the layers across different LLMs (the detailed results for probing formal validity and plausibility can be found in the supplementary material). Therefore, following the insights from probing, subsequent steering methods intervene within the third quarter of layers at the residual stream corresponding to the last input token position.

\paragraph{Evaluation metrics.} We adopt different evaluation metrics to compute the effect of steering on syllogistic reasoning. First, we compute the accuracy (ACC) of each model when assessing the validity of the syllogistic arguments in the test set. In addition, we measure the content effect (CE) based on the difference in accuracy on different subsets of the test set. In particular, we measure both the cross-plausibility CE as the difference in overall accuracy between plausible and implausible arguments, as well as the intra-plausibility CE as the difference in accuracy between valid and invalid arguments given a fixed plausibility value. The overall CE reported in the experiments is computed as the average of cross and intra-plausibility CE. Finally, we report the Acc/CE ratio, as the objective is to obtain maximal accuracy on formal reasoning with minimal content effect.

\paragraph{Computing steering vectors.} We compute the steering vectors following the methodology described in Section \ref{sec:method_steering}. In particular, we run each model on a training set composed of 2400 examples equally split across different types of arguments, and select as positive steering vectors the average of the activations that lead to correct predictions, and as negative steering vectors, the average of the activations that lead to wrong predictions.

\begin{table}[t]
  \centering
  \small
  \begin{tabular}{ll|ccc|c}
  \\
    \toprule
     \textbf{Model} & \textbf{Size} &  \textbf{Acc} $\uparrow$& \textbf{CE}  $\downarrow$& \textbf{Acc/CE} $\uparrow$& \textbf{ $\Delta_{rel}^{Acc/CE}$}\\
    \midrule
     Llama 3.2 & 3b & 77.79 & 17.50 & 4.45 & -\\
     Qwen 2.5 & 3b & \textbf{85.29}  & \textbf{7.12} & \textbf{11.99} & -\\
     \midrule
     \textbf{\cast{}}\\
     \midrule
     Llama 3.2 & 3b & 81.04 & 15.74 & 5.21 & 17.07 \\
     Qwen 2.5 & 3b & \textbf{85.86} & \textbf{4.42} & \textbf{19.41} & \textbf{61.88}\\
     \midrule
     \textbf{\ours{}}\\
     \midrule
     Llama 3.2 & 3b & \textbf{92.60} & \textbf{4.04} & \textbf{22.92} &  \textbf{415.05}\\
     Qwen 2.5 & 3b & 85.28 & 5.19 & 16.42 & 36.94\\  
     \bottomrule
     \\
  \end{tabular}
  
    \caption{Results of conditional steering on models that are unresponsive to static contrastive steering -- i.e, Llama 3.2 3b and Qwen 2.5 7b. We found that both \cast{} and \ours{} effectively improve Acc/CE.}
  \label{tab:conditional_results}
\end{table}

\begin{table*}[t!]
  \centering
  \small
  \begin{tabular}{ll|ccc|ccc|ccc}
    \toprule
    \textbf{Model} & \textbf{Size} &
    \multicolumn{3}{c|}{\textbf{English}} &
    \multicolumn{3}{c|}{\textbf{Chinese}} &
    \multicolumn{3}{c}{\textbf{German}} \\
    \cmidrule(lr){3-5} \cmidrule(lr){6-8} \cmidrule(lr){9-11}
    
    & & 
    
    $\textbf{PPL}_{\alpha=0}\downarrow$ & $\textbf{PPL}_{\alpha_{\text{best}}}\downarrow$ &
    $\Delta\%$ &
    
    $\textbf{PPL}_{\alpha=0}\downarrow$ & $\textbf{PPL}_{\alpha_{\text{best}}}\downarrow$  &  
    $\Delta\%$ &
    
    $\textbf{PPL}_{\alpha=0}\downarrow$ & $\textbf{PPL}_{\alpha_{\text{best}}}\downarrow$ &
    $\Delta\%$ \\
    \midrule
    Llama 3.2 & 1b & 24.29 & 24.77 & \textbf{1.98} & 49.81 & 51.35 & \textbf{3.09} & 20.16 & 20.52 & \textbf{1.79}\\
    Gemma 2   & 9b & 18.95 & 20.58 & 8.60 & 34.00 & 38.17 & 12.26 & 16.04 & 17.43 & 8.67\\
    Qwen 2.5  & 7b & \textbf{14.59} & \textbf{15.17} & 3.98 & \textbf{18.41} & \textbf{19.07} & 3.58 & \textbf{11.18} & \textbf{11.58} & 3.57\\
    \bottomrule \\
  \end{tabular}
    \begin{tabular}{ll|ccc|ccc}
    \toprule
    \textbf{Model} & \textbf{Size} &
    \multicolumn{3}{c|}{\textbf{ProntoQA}} &
    \multicolumn{3}{c}{\textbf{Rulebreakers}} \\
    \cmidrule(lr){3-5} \cmidrule(lr){6-8}
    & &
        $\textbf{ACC}_{\alpha=0}\uparrow$ & $\textbf{ACC}_{\alpha_{\text{best}}}\uparrow$ & 
        $\Delta\%$ & 
        
        $\textbf{ACC}_{\alpha=0}\uparrow$ & $\textbf{ACC}_{\alpha_{\text{best}}}\uparrow$ &
        $\Delta\%$
        \\
    \midrule
    Llama 3.2 & 1b & 49.6 & 53.6 & \textbf{8.1} & 40.2 & 38.6 & -4.0\\
    Gemma 2   & 9b & \textbf{62.2} & 52.2 & -16.1 & \textbf{92.0} & 85.6 & -6.9\\
    Qwen 2.5  & 7b & 53.6 & \textbf{56.4} & 5.2 & 88.2 & \textbf{88.2} & \textbf{0.0}\\
    \bottomrule
  \end{tabular}

\caption{(Top) Impact of steering on multilingual language modeling capabilities. (Bottom) Generalization to OOD logical reasoning tasks. The results demonstrate that steering for content effects incurs minimal side effects on multilingual language modeling capabilities, and can generalize to some extent to out-of-distribution reasoning tasks.}
  \label{tab:language_modeling_generalisation}
\end{table*}

\subsection{Contrastive Activation Steering}

We perform experiments on a test set of 2400 examples equally distributed across different syllogistic schemes. We investigate the effect of steering by varying the value of $\alpha$ between -3 and 3.  The results are reported in Table \ref{tab:res_static_steering}.

\paragraph{Effectiveness of contrastive steering.} The results reveal that contrastive steering is effective for improving Acc/CE on most of the tested models in both zero-shot and ICL settings. Notably, contrastive steering has the highest impact on Llama 3.2 1b with a relative improvement of Acc/CE of up to 777.27\%. A substantial improvement can be observed across different families and sizes of models (in particular for Gemma 2 9b and Qwen 2.5 7 via zero-shot). Moreover, we found that for most models, steering for content effect not only improves CE, but also contributes to significant improvements in accuracy on the syllogistic reasoning task (e.g., from 58.17\% to 73.56\% with Llama 1b). At the same time, steering with a static value of $\alpha$ is ineffective on two zero-shot models -- i.e., Llama 3.2 3b and Qwen 2.5 3b.

\paragraph{Steering outperforms ICL.} In general, we observe that ICL via few-shot examples is not sufficient to mitigate content effect biases and, in most cases, can have the detrimental effect of reducing accuracy. Contrastive steering, on the other hand, seems to be a much more effective methodology to mitigate reasoning biases in LLMs. This is confirmed by the fact that the best results on syllogistic reasoning are achieved by steered models.

\paragraph{The scaling parameter $\alpha$ enables explicit steering control.} In order to investigate why steering is not effective on Llama 3.2 3b and Qwen 2.5 3b, we study the detailed dynamics emerging with different values of $\alpha$. Here, we observe that, despite being ineffective on some models, contrastive steering can be used to explicitly control the accuracy achieved on valid and invalid arguments by varying the sign of $\alpha$. Specifically, the results show that setting $\alpha < 0$ generally improves accuracy on \textit{valid} arguments, while $\alpha > 0$ improves accuracy on \textit{invalid} arguments. This observation motivates us to explore conditional steering techniques to dynamically determine the value of $\alpha$ and attempt to steer unresponsive models.

\subsection{Conditional Activation Steering}

\paragraph{Computing Condition Vectors.} Motivated by the observation that the sign of $\alpha$ enables explicit control, we compute condition vectors to identify whether a given model is processing a valid or an invalid argument from the internal activations and then modulate the parameter $\alpha$ accordingly (i.e., setting $\alpha<0$ if \emph{valid}, and $\alpha>0$ otherwise). To this end, we collect condition activation vectors for validity from the training set and experiment with both \cast{} and \ours{}.

\paragraph{Conditional steering is effective on unresponsive models.} We found that both \cast{} and \ours{} are effective in improving Acc/CE for both Llama 3b and Qwen 3b (see Table \ref{tab:conditional_results}). Moreover, while the results on Qwen show that \cast{} and \ours{} have a similar effect, the results on Llama reveal that \ours{} is significantly more effective, leading to an absolute increase in accuracy of up to 15\%. Figure \ref{fig:detailed_contrastive_steering_main} (b) shows the impact of \ours{} on Llama 3b with different values of $\alpha$ (with the sign dynamically determined).

\begin{figure}[t]
\centering
\includegraphics[width=\columnwidth]{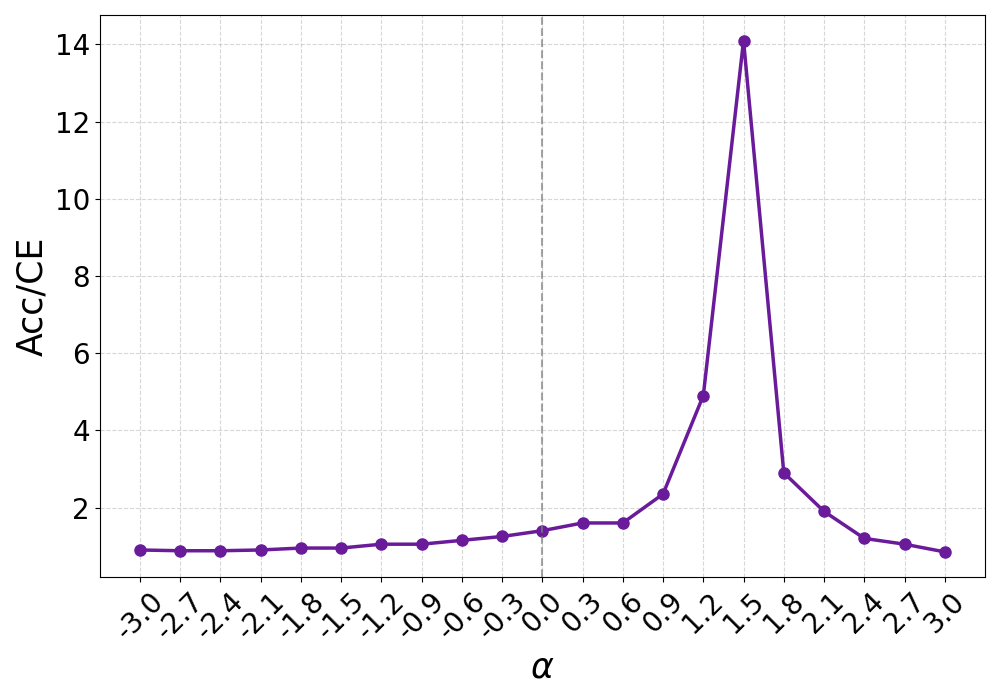}
\caption{Robustness of steering to prompt variations on Llama 1b (i.e. Acc/CE). The results reveal that, despite specific variations deriving from perturbations applied at test time, the overall effectiveness of steering remains unaltered.}
\label{fig:robustness_llama}
\end{figure}

\subsection{Robustness to Prompt Perturbations} 

To test the robustness of steering performance to prompt perturbations~\cite{mizrahi-etal-2024-state}, we construct a set of prompt variants by employing instruction templates different from those used in the training set (i.e. instruction template paraphrasing). Following~\cite{mizrahi-etal-2024-state}, we employ two prompting strategies proven effective in prior research:
(1) \textit{Instruction template rephrasing}: we use GPT-4.5 to paraphrase a seed instruction template~\cite{lester2021power,gonen2022demystifying, honovich-etal-2023-unnatural};
(2) \textit{Instruction induction}: inspired by~\cite{honovich-etal-2023-instruction}, we provide five input-output pairs and ask GPT-4.5 to generate the possible instructions. Given a set of prompt variations (see supplementary material), we compute the steering vectors using the original prompt and randomly select a variant at inference time. 

\paragraph{Steering is robust to prompt variations.} The results in Figure \ref{fig:robustness_llama} on Llama 1b (i.e., the model with the best Acc/CE improvement on the original prompt) reveal that, despite specific variation in the values of $\alpha_{best}$ and some noise deriving from prompt variation, the overall effectiveness of steering remains unaltered. A similar trend is also observable for other models (see supplementary material).

\subsection{Impact on Non-Target Capabilities}

Ideally, the steering effect should be localised -- i.e., do not impact non-target capabilities. In this section, we particularly consider the language modeling capability and the reasoning capability on out-of-distribution (OOD) tasks: information-informed reasoning and multi-premise deductive reasoning. For each experiment, we compare the performance between the model without steering ($\alpha=0$ in Table~\ref{tab:language_modeling_generalisation}) against the model with steering ($\alpha_{best}$). If steering is perfectly localized, the two should perform indistinguishably on these non-target tasks. 

\paragraph{Multilingual language modeling.} We draw 2,000 examples per language from the C4 dataset \cite{raffel2020exploring} and compute the average perplexity on causal language modeling  over sequences of length 1,024. Table~\ref{tab:language_modeling_generalisation} shows that content-effect steering leaves multilingual modeling nearly intact. For example, on English text, Llama 3.2 1b model’s perplexity changes only from 24.29 (baseline) to 24.77 (steered). Gemma exhibits the most significant relative increase, but even there, the gap remains small; across all languages and model sizes, perplexity deviations stay within a few percent.

\paragraph{OOD reasoning tasks.} 
We test the steering impact on two reasoning tasks (i.e., ProntoQA \cite{PrOntoQA} and Rulebreakers \cite{chan2024rulebreakers}) that were not presented during the steering modulating process. ProntoQA is a synthetic task designed to test deductive reasoning on multiple natural language premises. Rulebreakers is a task for evaluating LLMs ability to distinguish whether logical entailment diverges from factually acceptable inference. Table~\ref{tab:language_modeling_generalisation} 
show that adopting the steering vectors computed on syllogisms generalise well, especially on ProntoQA with Llama and Qwen (+8.1\% and +5.2\%), whereas Gemma experiences a substantial performance drop on both tasks, most notably a 16.1\% decrease on ProntoQA (this aligns with the higher drops observed on language modeling). 

These findings underscore both the promise of steering for enhancing targeted reasoning and the persistent challenge of achieving a complete and robust OOD generalization.


\section{Conclusion}

This paper investigated how to mitigate content effects in LLMs' reasoning. We systematically tested static and conditional activation steering techniques to disentangle formal validity from content plausibility in syllogistic reasoning. 
Our results indicate that steering is particularly effective in reducing content effect and improving accuracy in formal reasoning. 
 In general,  this paper demonstrates that activation-level interventions offer a scalable inference-time strategy and can contribute towards more systematic and unbiased formal reasoning in LLMs. 


\section*{Acknowledgments}
This work was partially funded by the SNSF project NeuMath (200021\_204617), by the CRUK National Biomarker Centre, and supported by the Manchester Experimental Cancer Medicine Centre and the NIHR Manchester Biomedical Research Centre.

\bibliography{aaai2026}

\appendix
\section{Experimental Setup}

We run all the steering experiments on a single A100 GPU. For probing, the graphs report the variance obtained with different random seeds (across 10 runs). For steering, We set the seed to 0 to enable a reproducible behaviour.

\section{Syllogistic Schemes}

Here is the list of syllogistic schemes adopted to construct the dataset. The schemes are aligned with the ones used in previous work on syllogistic reasoning with LLMs \cite{bertolazzi-etal-2024-systematic}.

\begin{lstlisting}[numbers=none]

Schema: AA1
Premise 1: All <A> are <B>
Premise 2: All <B> are <C>
Conclusions: All <A> are <C> | some <A> are <C> | some <C> are <A>
 
Schema: AA2
Premise 1: All <B> are <A>
Premise 2: All <C> are <B>
Conclusions: All <C> are <A> | some <A> are <C> | some <C> are <A>
 
Schema: AA4
Premise 1: All <B> are <A>
Premise 2: All <B> are <C>
Conclusions: Some <A> are <C> | some <C> are <A>
 
Schema: AI2
Premise 1: All <B> are <A>
Premise 2: Some <C> are <B>
Conclusions: Some <A> are <C> | some <C> are <A>
 
Schema: AI4
Premise 1: All <B> are <A>
Premise 2: Some <B> are <C>
Conclusions: Some <A> are <C> | some <C> are <A>
 
Schema: AO3
Premise 1: All <A> are <B>
Premise 2: Some <C> are not <B>
Conclusions: Some <C> are not <A>
 
Schema: AO4
Premise 1: All <B> are <A>
Premise 2: Some <B> are not <C>
Conclusions: Some <A> are not <C>
 
Schema: AE1
Premise 1: All <A> are <B>
Premise 2: No <B> are <C>
Conclusions: No <A> are <C> | no <C> are <A> | some <A> are not <C> | some <C> are not <A>
 
Schema: AE2
Premise 1: All <B> are <A>
Premise 2: No <C> are <B>
Conclusions: Some <A> are not <C>
 
Schema: AE3
Premise 1: All <A> are <B>
Premise 2: No <C> are <B>
Conclusions: No <A> are <C> | no <C> are <A> | some <A> are not <C> | some <C> are not <A>
 
Schema: AE4
Premise 1: All <B> are <A>
Premise 2: No <B> are <C>
Conclusions: Some <A> are not <C>
 
Schema: IA1
Premise 1: Some <A> are <B>
Premise 2: All <B> are <C>
Conclusions: Some <A> are <C> | some <C> are <A>
 
Schema: IA4
Premise 1: Some <B> are <A>
Premise 2: All <B> are <C>
Conclusions: Some <A> are <C> | some <C> are <A>
 
Schema: IE1
Premise 1: Some <A> are <B>
Premise 2: No <B> are <C>
Conclusions: Some <A> are not <C>
 
Schema: IE2
Premise 1: Some <B> are <A>
Premise 2: No <C> are <B>
Conclusions: Some <A> are not <C>
 
Schema: IE3
Premise 1: Some <A> are <B>
Premise 2: No <C> are <B>
Conclusions: Some <A> are not <C>
 
Schema: IE4
Premise 1: Some <B> are <A>
Premise 2: No <B> are <C>
Conclusions: Some <A> are not <C>
 
Schema: OA3
Premise 1: Some <A> are not <B>
Premise 2: All <C> are <B>
Conclusions: Some <A> are not <C>
 
Schema: OA4
Premise 1: Some <B> are not <A>
Premise 2: All <B> are <C>
Conclusions: Some <C> are not <A>
 
Schema: EA1
Premise 1: No <A> are <B>
Premise 2: All <B> are <C>
Conclusions: Some <C> are not <A>
 
Schema: EA2
Premise 1: No <B> are <A>
Premise 2: All <C> are <B>
Conclusions: No <A> are <C> | no <C> are <A> | some <A> are not <C> | some <C> are not <A>
 
Schema: EA3
Premise 1: No <A> are <B>
Premise 2: All <C> are <B>
Conclusions: No <A> are <C> | no <C> are <A> |some <A> are not <C> | some <C> are not <A>
 
Schema: EA4
Premise 1: No <B> are <A>
Premise 2: All <B> are <C>
Conclusions: Some <C> are not <A>
 
Schema: EI1
Premise 1: No <A> are <B>
Premise 2: Some <B> are <C>
Conclusions: Some <C> are not <A>
 
Schema: EI2
Premise 1: No <B> are <A>
Premise 2: Some <C> are <B>
Conclusions: Some <C> are not <A>
 
Schema: EI3
Premise 1: No <A> are <B>
Premise 2: Some <C> are <B>
Conclusions: Some <C> are not <A>
 
Schema: EI4
Premise 1: No <B> are <A>
Premise 2: Some <B> are <C>
Conclusions: Some <C> are not <A>
\end{lstlisting}

\section{Prompts Templates}

\begin{lstlisting}[numbers=none]

Variation 0:

Given the premises, evaluate 
the validity of the conclusion.

Premises:
entry[`premise1'].
entry[`premise2'].

Conclusion: entry[`conclusion'].

The conclusion is

Variation 1:

Carefully evaluate the validity
of the following logical argument and
answer'the argument is logically valid' 
or 'the argument is logically invalid'.

Premise1: entry[`premise1'].
Premise2: entry[`premise2'].
Conclusion: entry[`conclusion'].

The argument is logically

Variation 2:

Analyze the formal logical structure of 
the argument below, then indicate 
whether it is valid or invalid.

Premise 1: entry[`premise1']\n
Premise 2: entry[`premise2']\n
Conclusion: entry[`conclusion']\n\n

The logical structure is

Variation 3:

Assess carefully whether the argument 
presented below is logically valid or 
invalid based on the given premises and 
conclusion.
 
1. entry[`premise1']
2. entry[`premise2']

Conclusion:
entry[`conclusion']

The argument is logically

Variation 4:

Examine the following argument, 
generating "valid" if the conclusion
logically follows from the premises 
provided, "invalid" otherwise.
 
- Premise 1: entry[`premise1']
- Premise 2: entry[`premise2']
Conclusion: entry[`conclusion']

The argument is

Variation 5:

Given the two premises and the 
conclusion below, judge carefully 
whether the logical argument is valid or 
invalid.

Premise 1: entry[`premise1']
Premise 2: entry[`premise2']
Conclusion: entry[`conclusion']

The logical argument is

Variation 6:

Evaluate the following logical argument
carefully and decide whether the 
provided conclusion is formally valid or 
invalid given the two premises.

Premise 1: entry[`premise1']
Premise 2: entry[`premise2']
Conclusion: entry[`conclusion']

The conclusion is formally
\end{lstlisting}

\section{Linear Probing Results}

\begin{figure*}
\centering

\subfloat[Plausibility -- Llama \label{fig:prob_pls_llama}]{\includegraphics[width=0.5\textwidth]{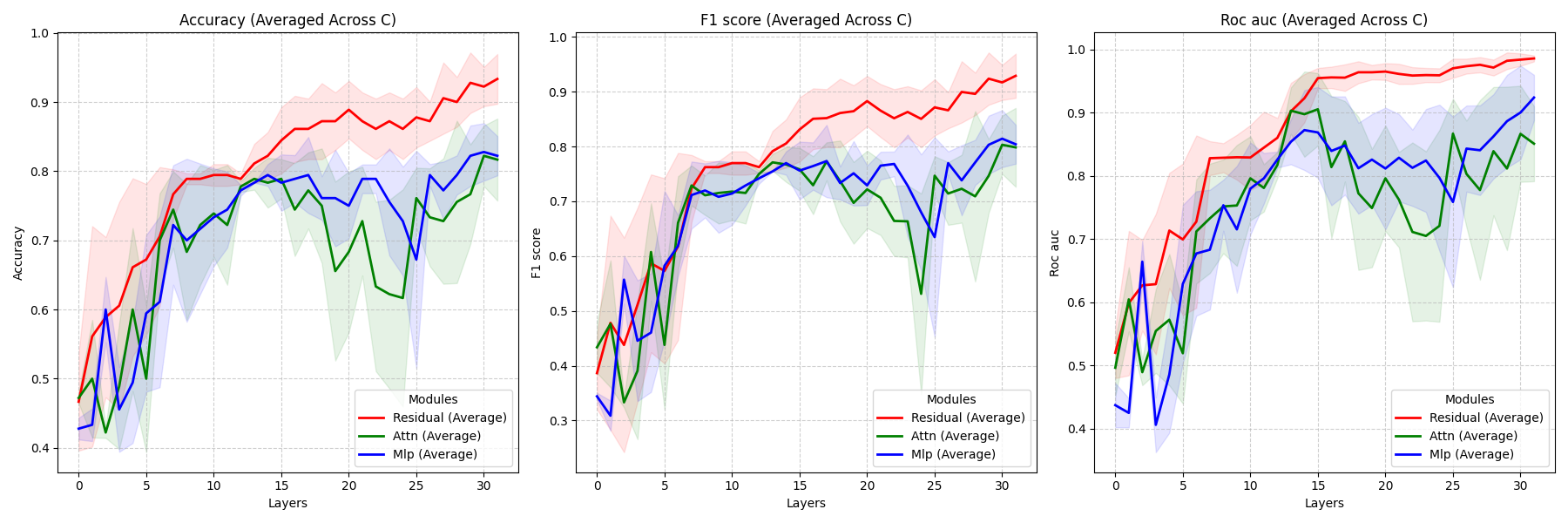}}

\subfloat[Validity -- Gemma 9b\label{fig:prob_val_gemma}]{\includegraphics[width=0.5\textwidth]{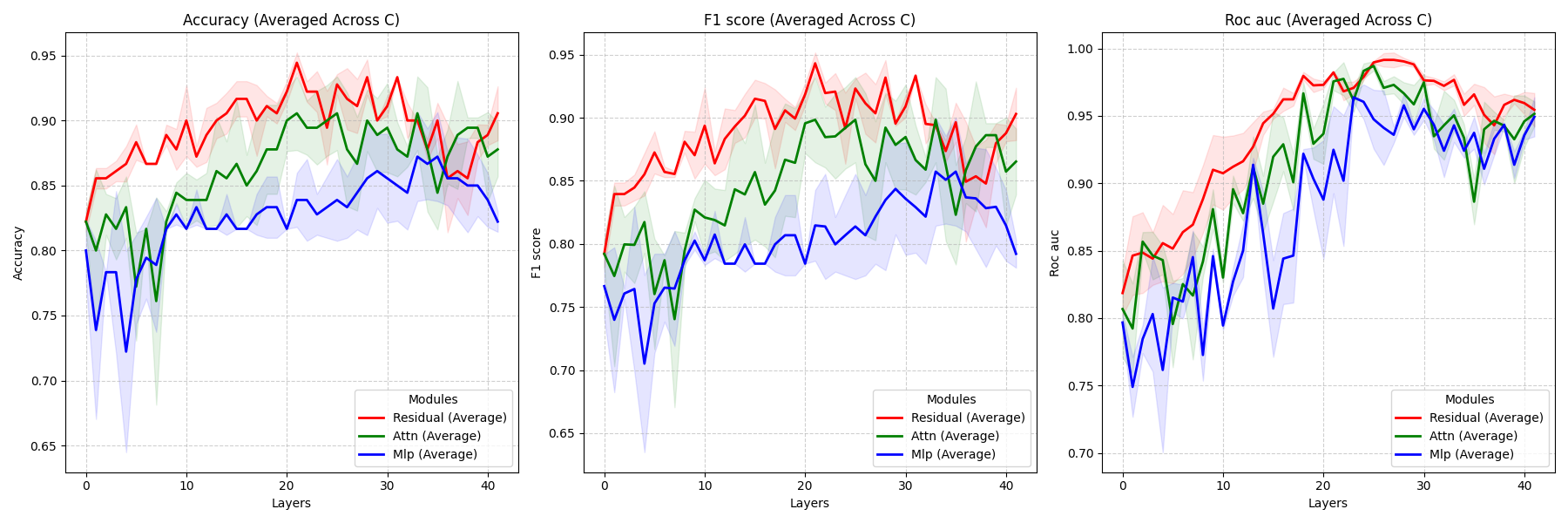}}

\subfloat[Plausibility -- Gemma 9b\label{fig:prob_pls_gemma}]{\includegraphics[width=0.5\textwidth]{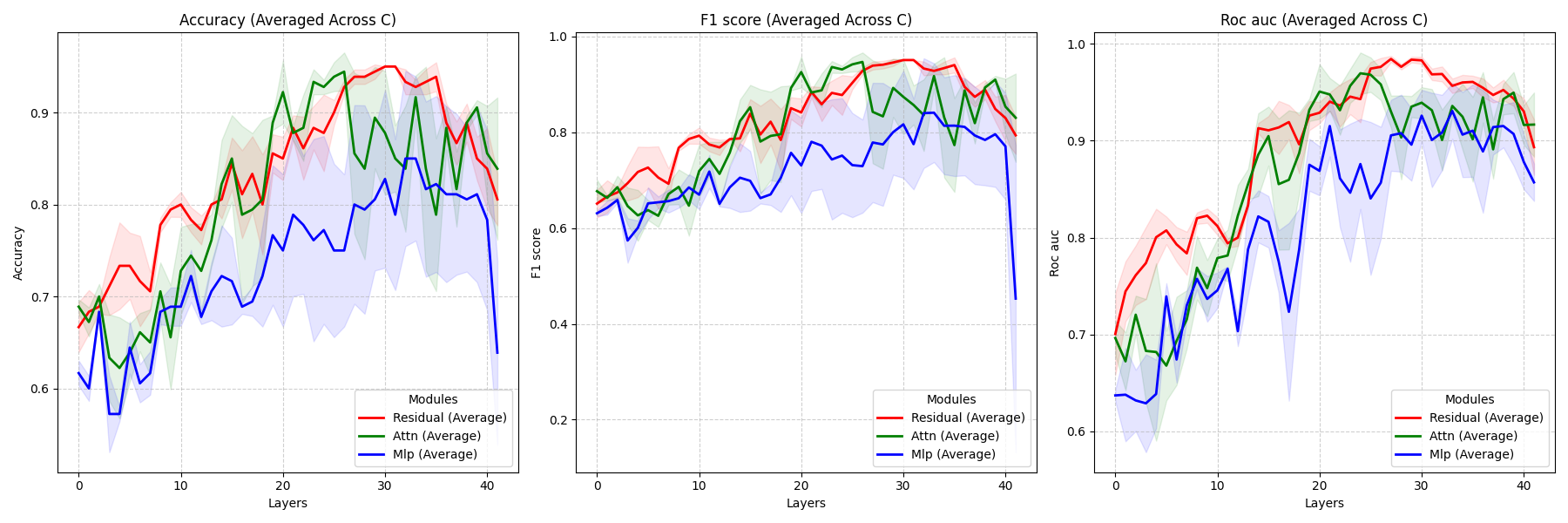}}

\subfloat[Validity -- Qwen 7b\label{fig:prob_val_qwen}]{\includegraphics[width=0.5\textwidth]{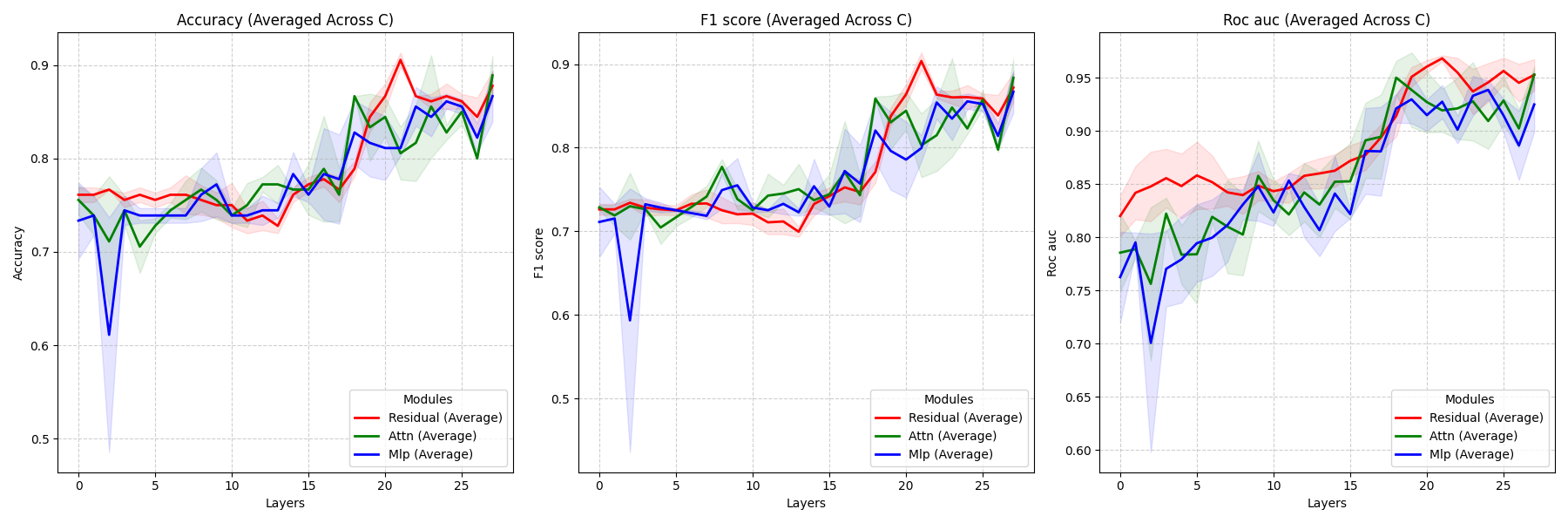}}

\subfloat[Validity -- Qwen 7b\label{fig:prob_pls_qwen}]{\includegraphics[width=0.5\textwidth]{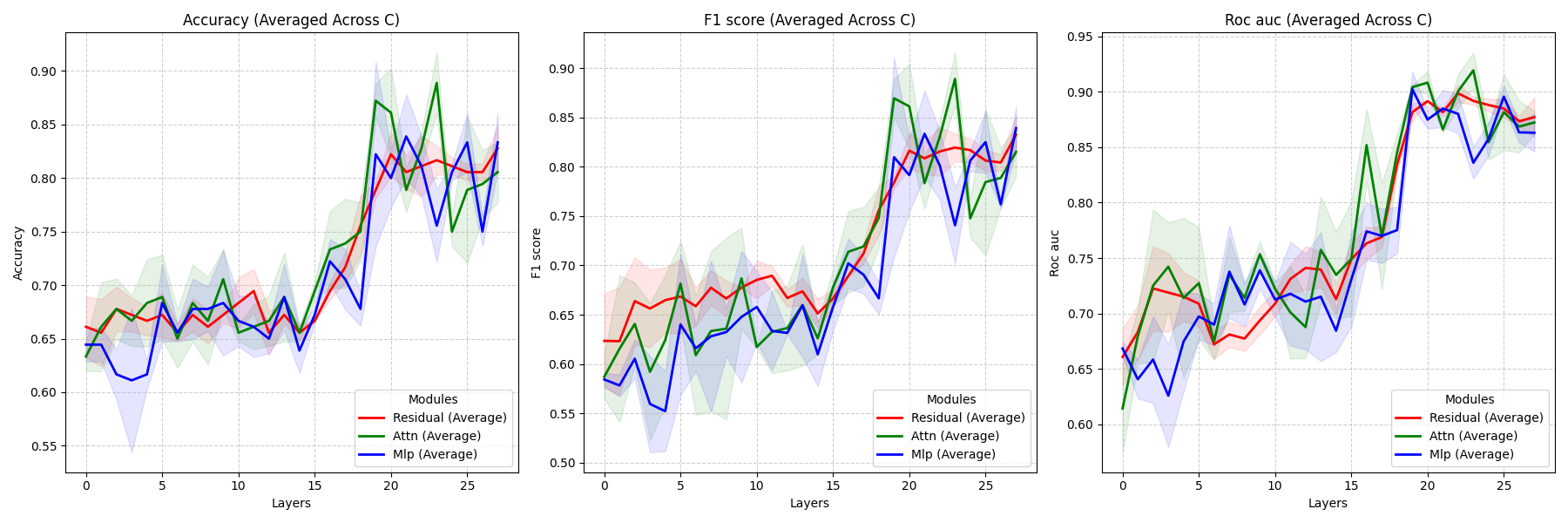}}

\caption{Linear Probing Results for different models. In general, the probing results suggest that the information about validity and plausibility is encoded in the second half of the layers, with a peak in the third quarter (predominantly in the residual stream).}
\label{fig:complete_probing}
\end{figure*}

Results are reported in Figure \ref{fig:probing} and Figure \ref{fig:complete_probing}.

\section{Static and Conditional Steering}

\begin{figure*}[t]
\centering
\subfloat[Acc -- Llama 1b\label{fig:acc_llama_1b}]{\includegraphics[width=0.32\textwidth]{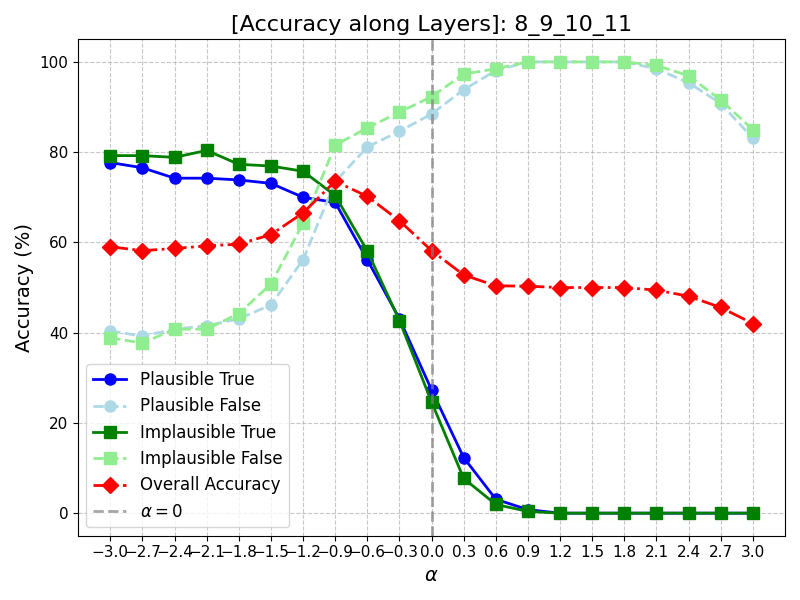}}
\subfloat[CE -- Llama 1b\label{fig:ce_llama_1b}]{\includegraphics[width=0.32\textwidth]{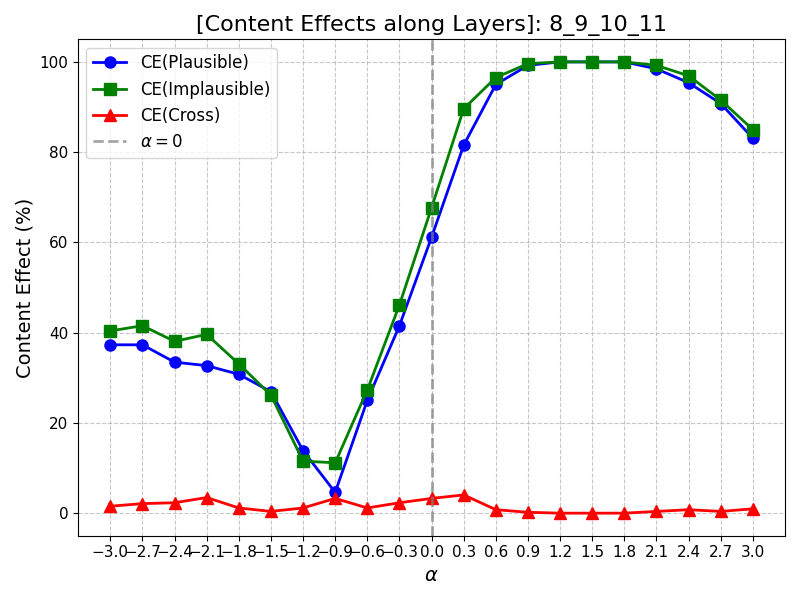}}
\subfloat[Acc/CE -- Llama 1b\label{fig:acc_ce_llama_1b}]{\includegraphics[width=0.32\textwidth]
{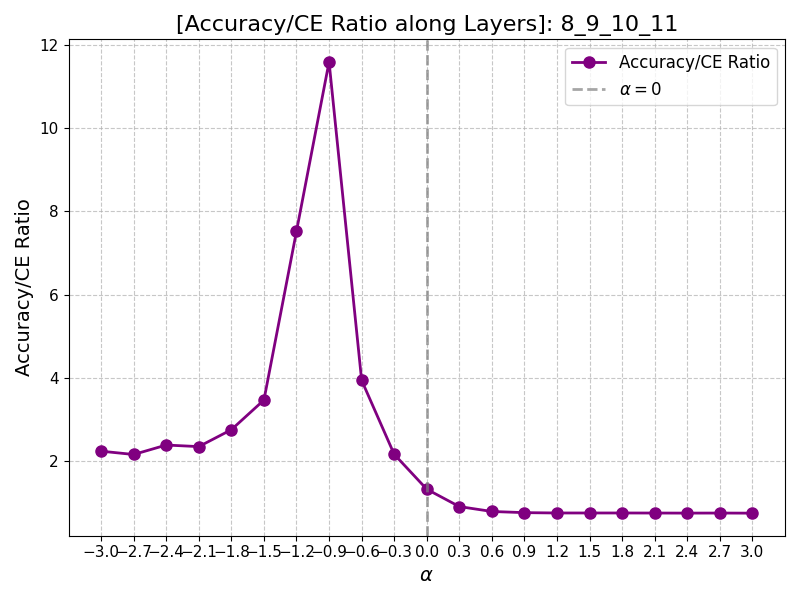}}

\subfloat[Acc -- Llama 3b \label{fig:acc_llama_3b}]{\includegraphics[width=0.32\textwidth]{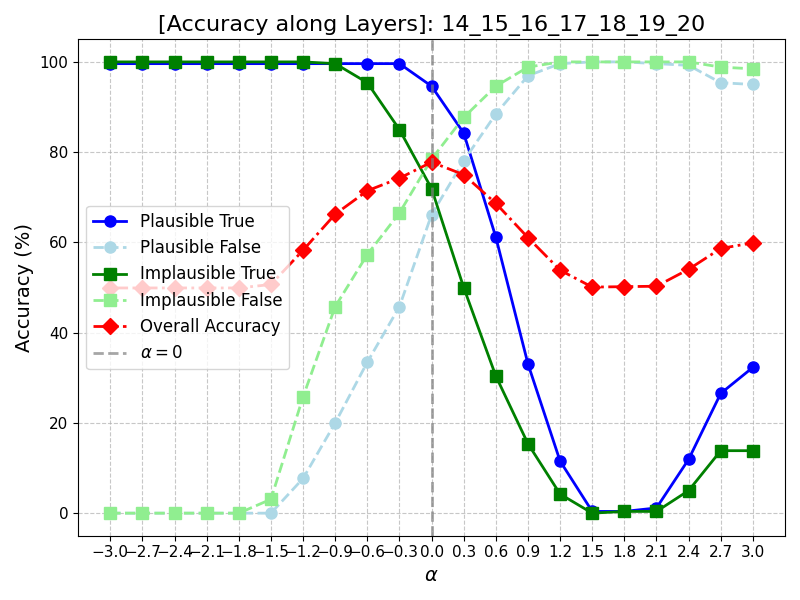}}
\subfloat[CE -- Llama 3b\label{fig:acc_llama_3b}]{\includegraphics[width=0.32\textwidth]{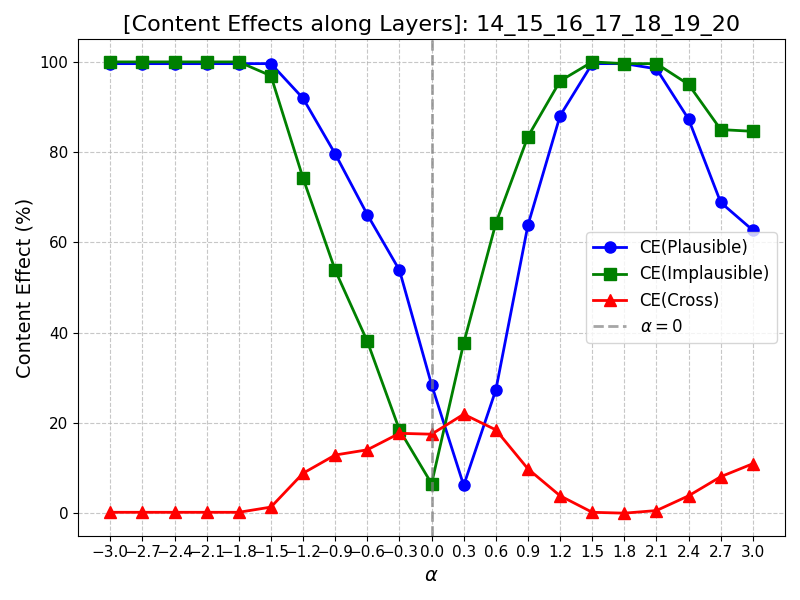}}
\subfloat[Acc/CE -- Llama 3b\label{fig:acc_ce_llama_3b}]{\includegraphics[width=0.32\textwidth]
{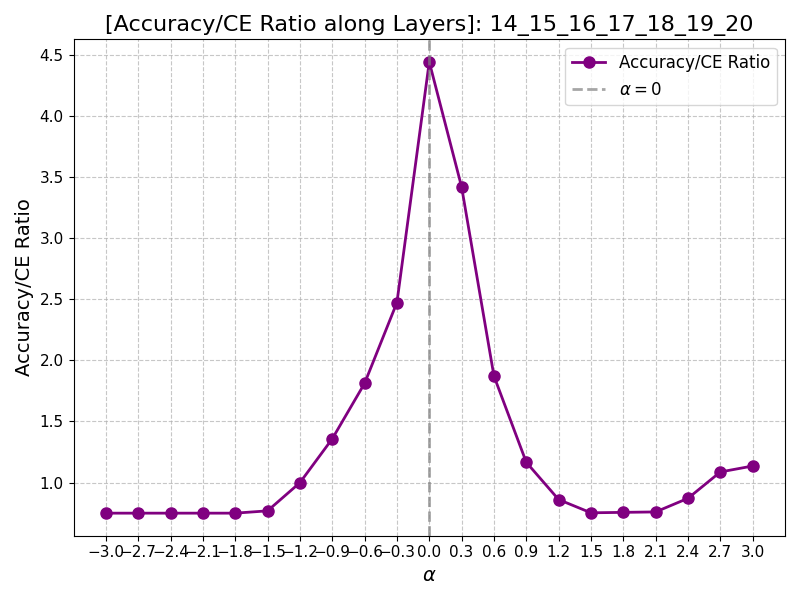}}

\subfloat[Acc -- \ours\label{fig:acc_knn}]{\includegraphics[width=0.32\textwidth]{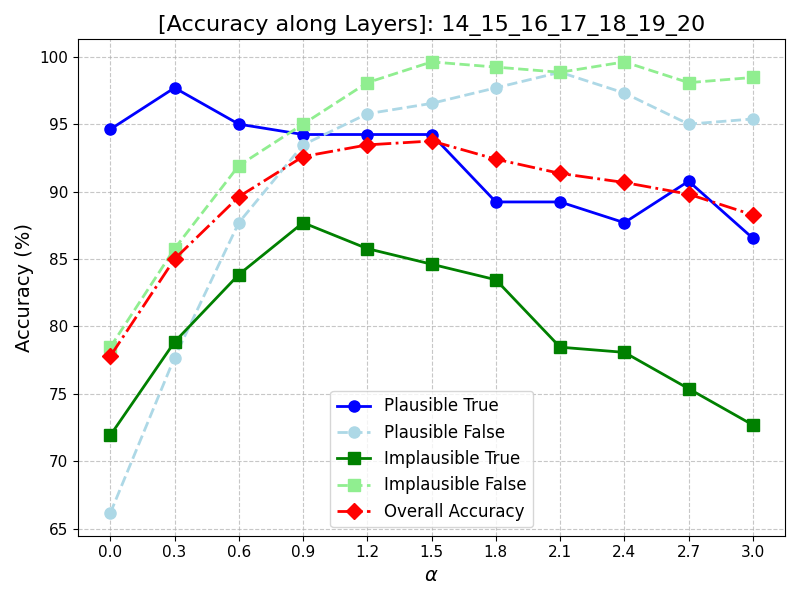}}
\subfloat[CE -- \ours \label{fig:ce_knn}]{\includegraphics[width=0.32\textwidth]{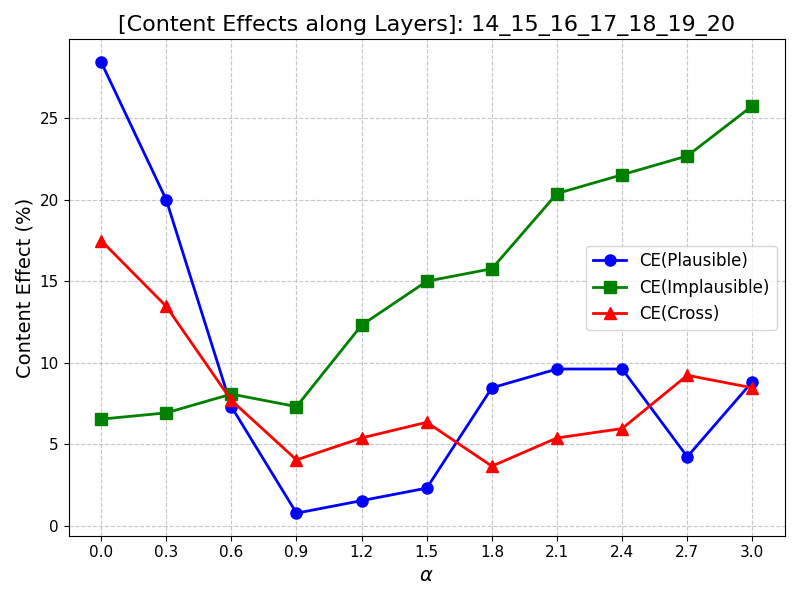}}
\subfloat[Acc/CE -- \ours \label{fig:acc_ce_knn}]{\includegraphics[width=0.32\textwidth]{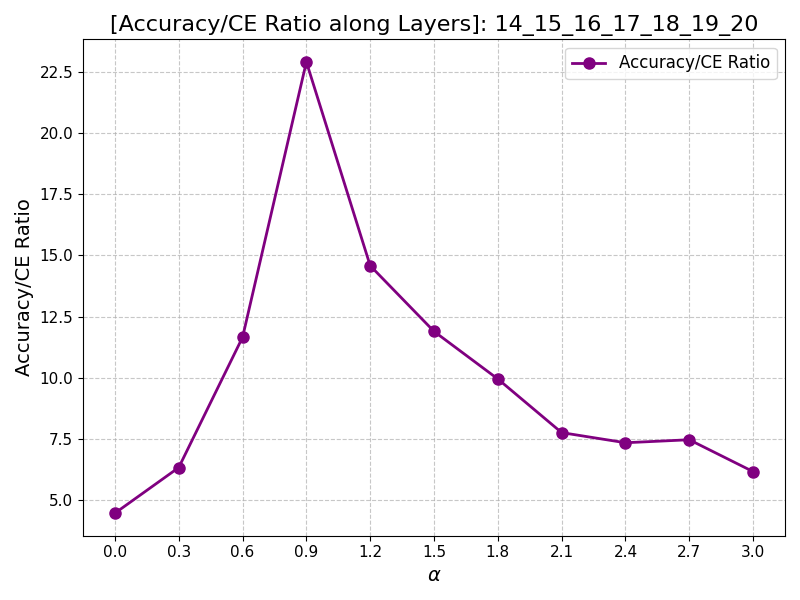}}

\caption{(Top) Example of effective (Llama 1b) and ineffective (Llama 3b) contrastive steering with static values of $\alpha$. (Bottom) Impact of conditional activation steering (\ours{}) on Llama 3b. Contrary to static steering, \ours{} leads to a significant increase in accuracy for Llama 3b (i.e., up to 15\%) while substantially reducing content effect.}
\label{fig:detailed_contrastive_steering}
\end{figure*}

\begin{figure*}[h]
\centering
\subfloat[Acc -- \cast{} \label{fig:acc_cond}]{\includegraphics[width=0.32\textwidth]{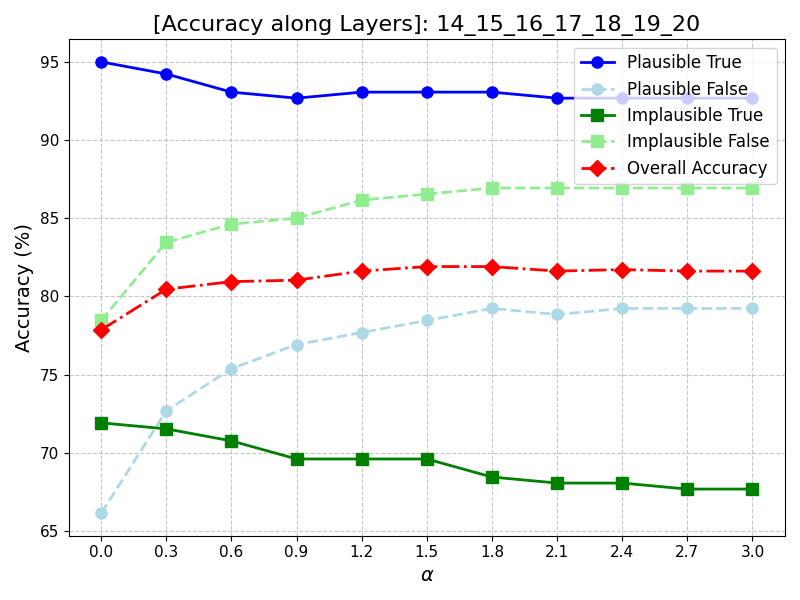}}
\subfloat[CE -- \cast{} \label{fig:ce_cond}]{\includegraphics[width=0.32\textwidth]{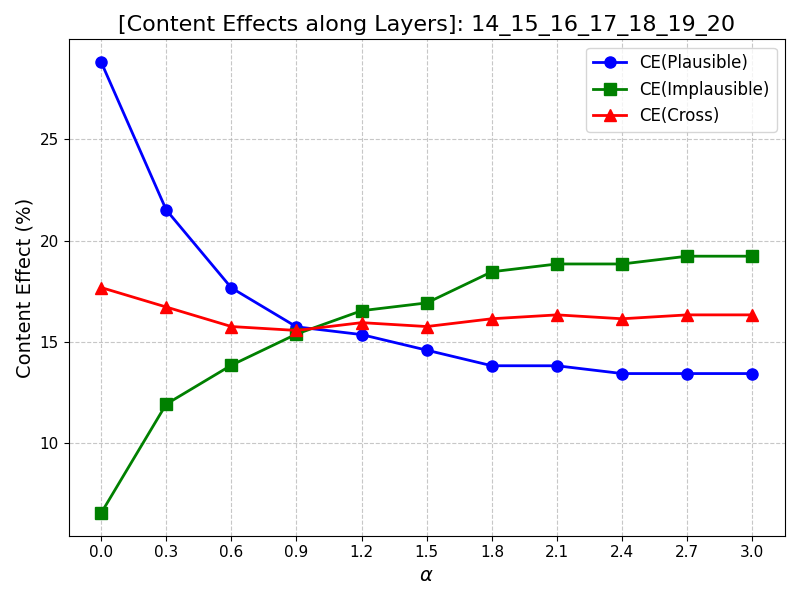}}
\subfloat[Acc/CE -- \cast{} \label{fig:acc_ce_cond}]{\includegraphics[width=0.32\textwidth]{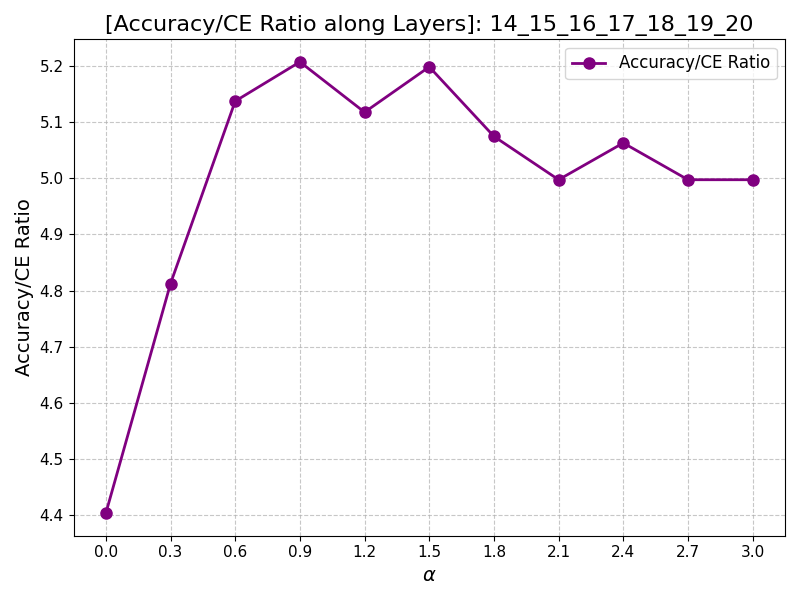}}

\subfloat[Acc -- \ours \label{fig:acc_knn}]{\includegraphics[width=0.32\textwidth]{figures/acc_knn_llama_3b.png}}
\subfloat[CE -- \ours \label{fig:ce_knn}]{\includegraphics[width=0.32\textwidth]{figures/ce_knn_llama_3b.png}}
\subfloat[Acc/CE -- \ours \label{fig:acc_ce_knn}]{\includegraphics[width=0.32\textwidth]{figures/acc_ce_knn_llama_3b.png}}

\caption{Results of conditional activation steering on Llama-3.2-3b-Instruct. Standard conditional steering (top) and KNN-based conditional steering (bottom).}
\label{fig:cond_steering_comparison}
\end{figure*}

\begin{figure*}[h]
\centering
\subfloat[Acc -- \cast{} \label{fig:acc_cond}]{\includegraphics[width=0.32\textwidth]{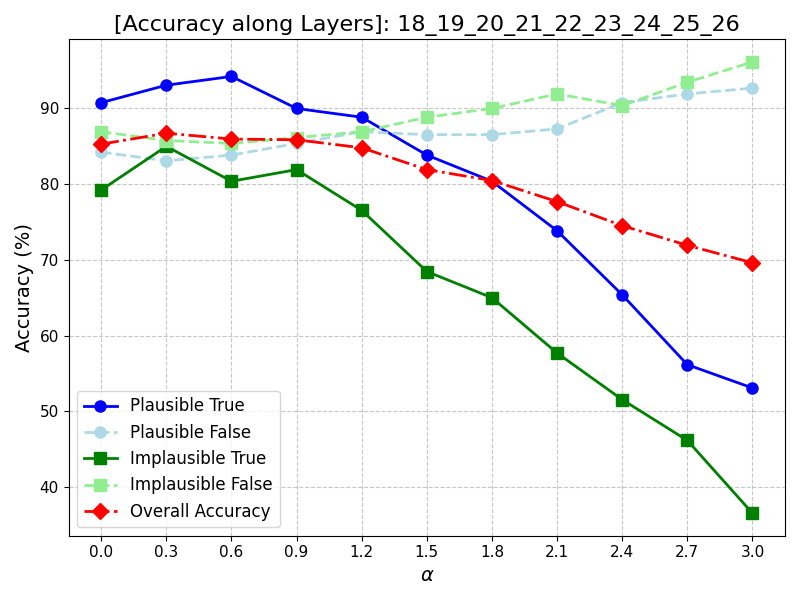}}
\subfloat[CE -- \cast{} \label{fig:ce_cond}]{\includegraphics[width=0.32\textwidth]{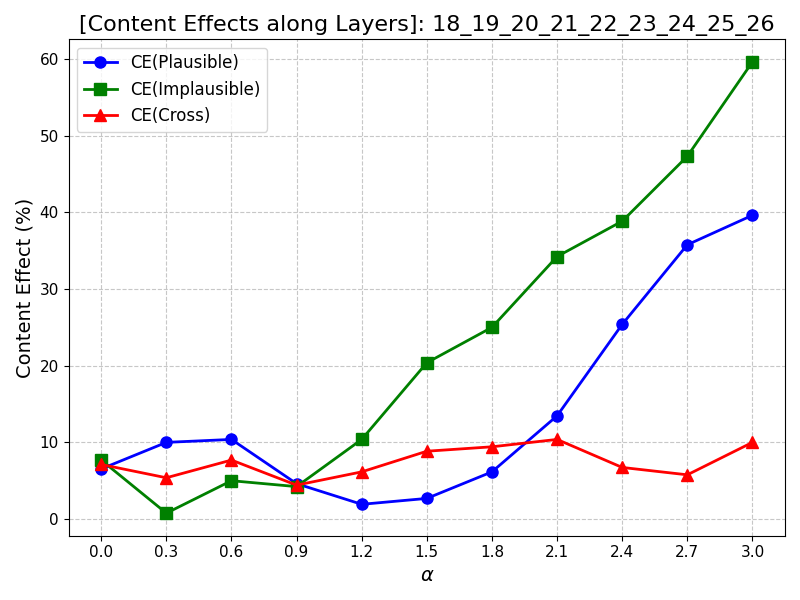}}
\subfloat[Acc/CE -- \cast{} \label{fig:acc_ce_cond}]{\includegraphics[width=0.32\textwidth]{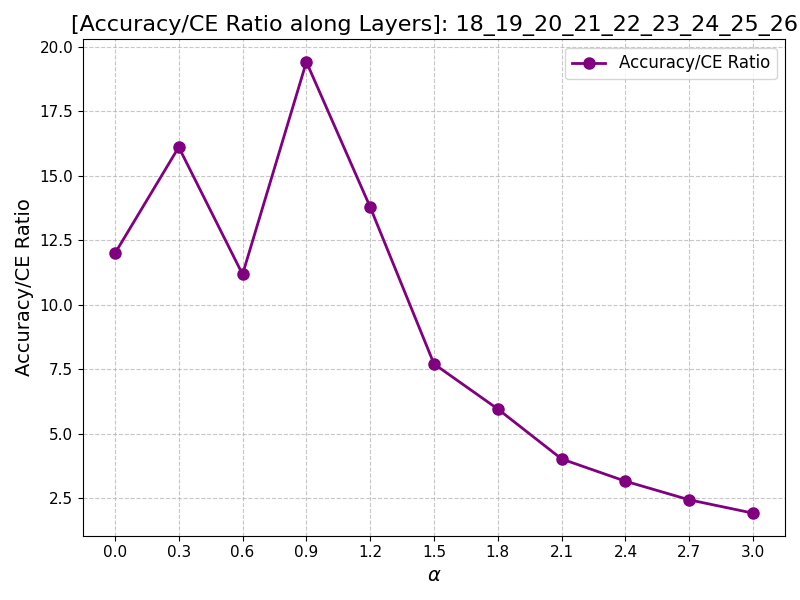}}

\subfloat[Acc -- \ours \label{fig:acc_knn}]{\includegraphics[width=0.32\textwidth]{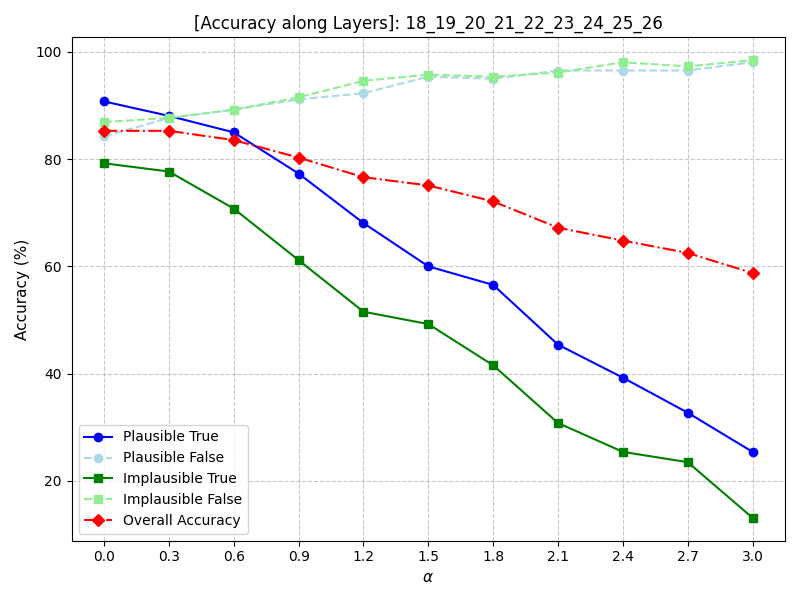}}
\subfloat[CE -- \ours \label{fig:ce_knn}]{\includegraphics[width=0.32\textwidth]{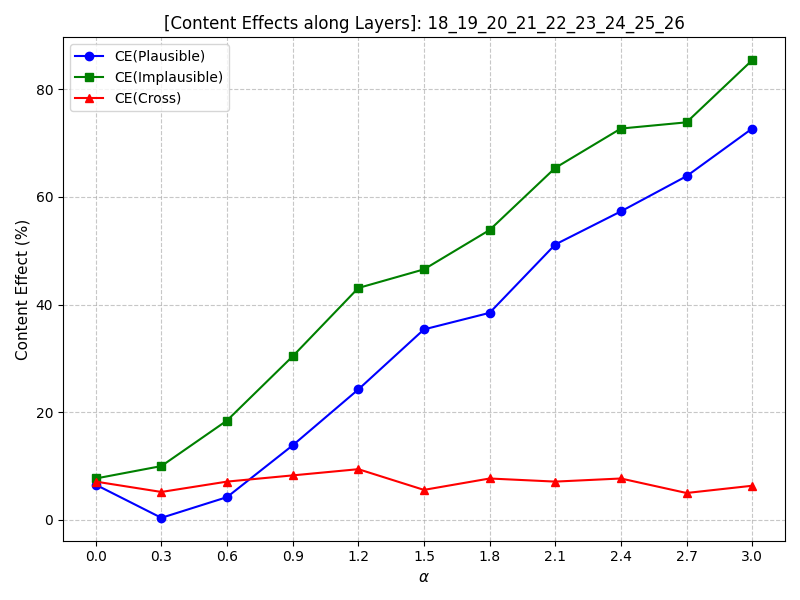}}
\subfloat[Acc/CE -- \ours \label{fig:acc_ce_knn}]{\includegraphics[width=0.32\textwidth]{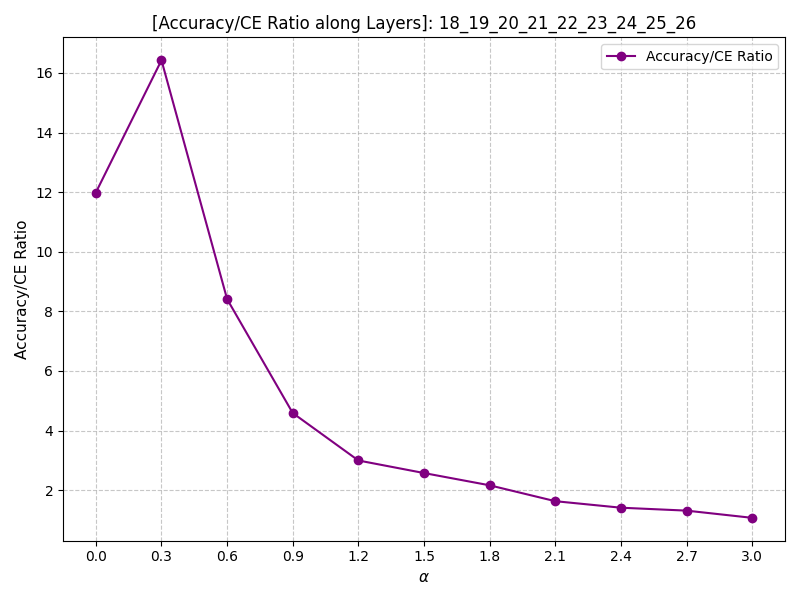}}

\caption{Results of conditional activation steering on Qwen 2.5 3b. Standard conditional steering (top) and KNN-based conditional steering (bottom).}
\label{fig:cond_steering_comparison_qwen}
\end{figure*}

Detailed results of static and conditional steering on Llama are reported in Figure \ref{fig:detailed_contrastive_steering}. A comparison between \cast{} and \ours{} is reported in Figure \ref{fig:cond_steering_comparison} and Figure \ref{fig:cond_steering_comparison_qwen}.

\section{Robustness to Prompt Perturbations}

\begin{figure*}[t]
\centering
\includegraphics[width=0.32\textwidth]{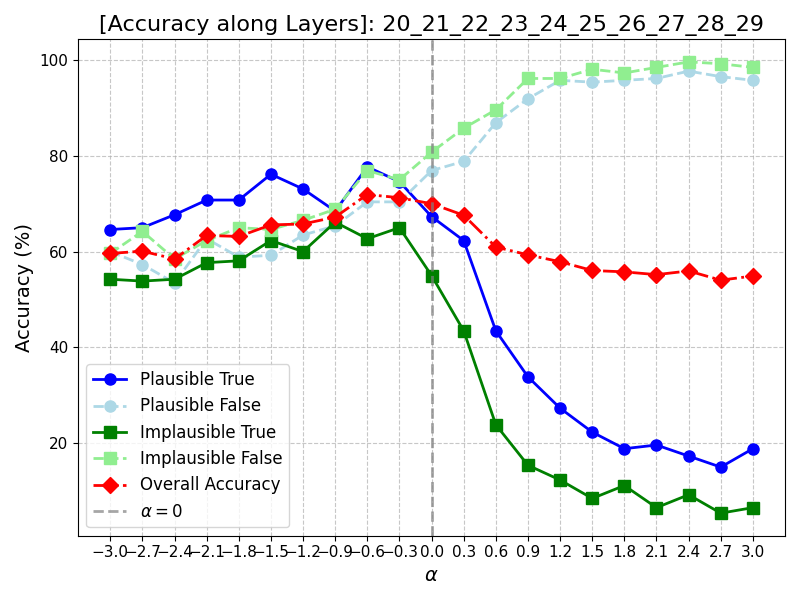}
\includegraphics[width=0.32\textwidth]{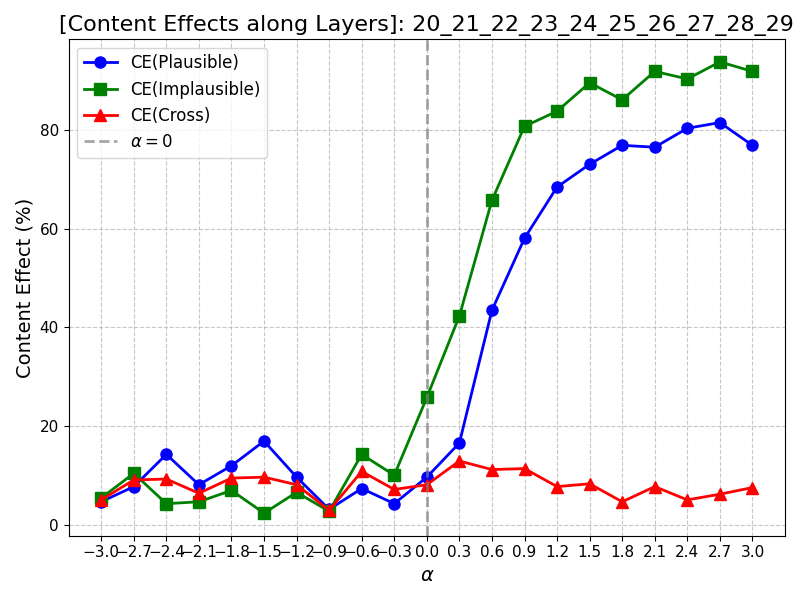}
\includegraphics[width=0.32\textwidth]{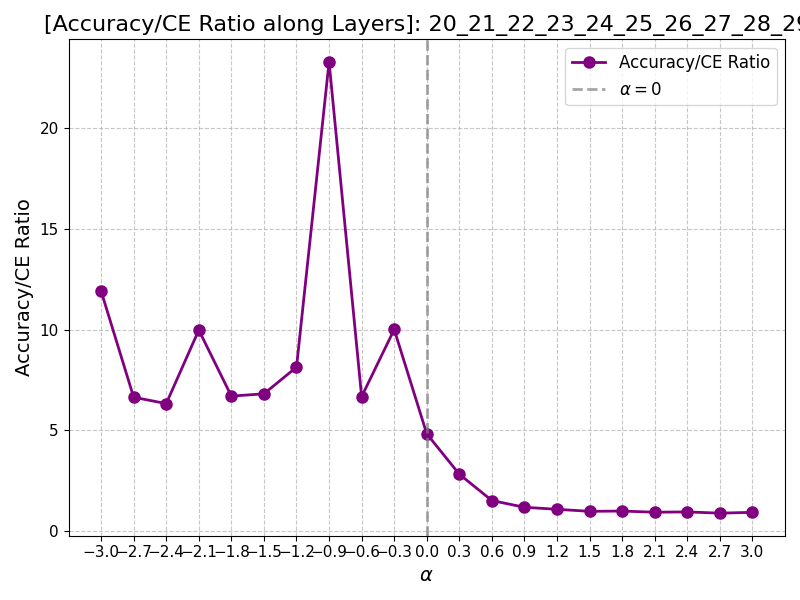}
\includegraphics[width=0.32\textwidth]{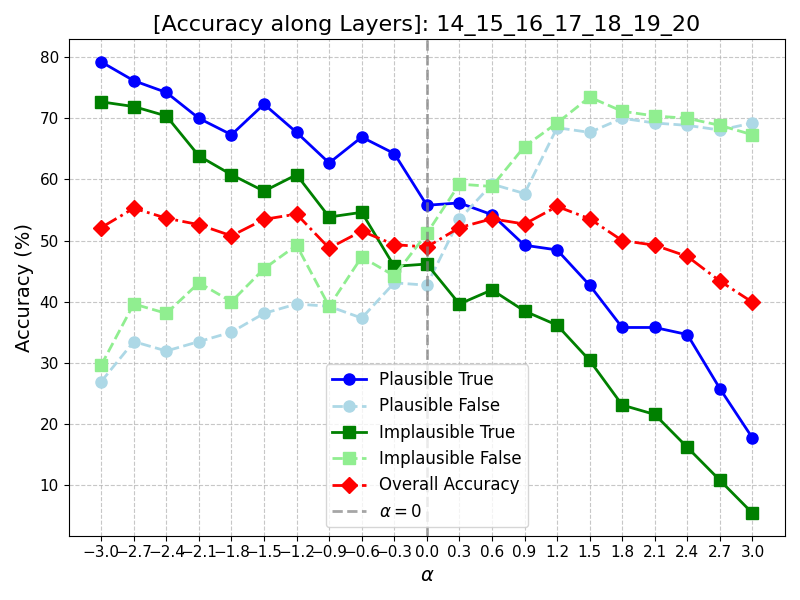}
\includegraphics[width=0.32\textwidth]{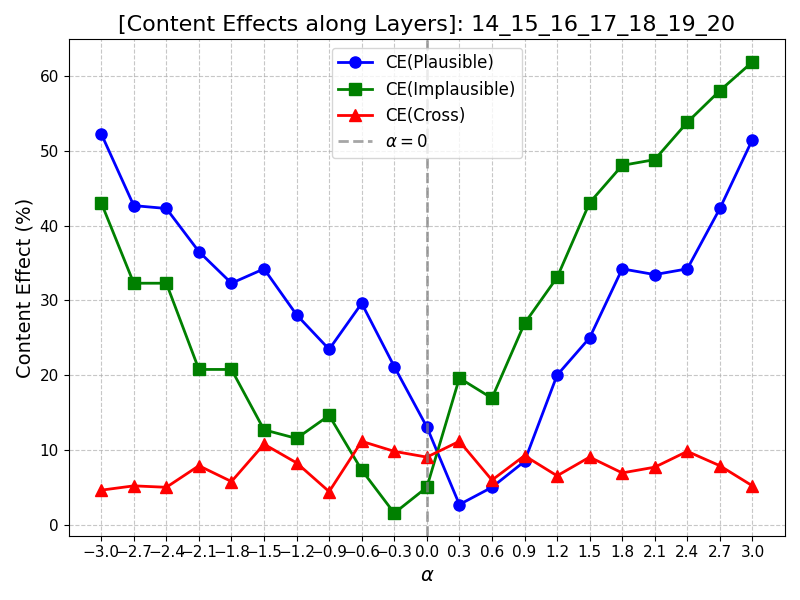}
\includegraphics[width=0.32\textwidth]{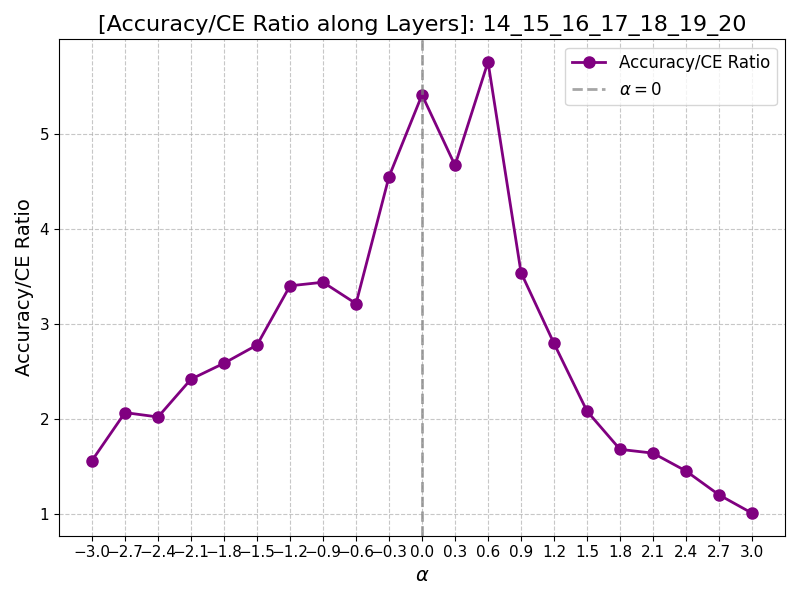}
\caption{Robustness of steering to prompt perturbations on Gemma 2 9b (top) and Qwen 2.5 7b (bottom) (i.e., ACC, CE, and ACC/CE from left to right). Steering is still effective despite variations applied to the prompts at test time.}
\label{fig:robustness_gemma_qwen}
\end{figure*}

Detailed results are reported in Figure \ref{fig:robustness_llama} and Figure \ref{fig:robustness_gemma_qwen}.

\section{Limitations}

We acknowledge existing limitations with the current study:

The experiments are performed on LLMs of up to a size of 9 billion due to computational constraints. While the framework introduced in this paper is general and can in principle be applied to larger models, we left such an investigation to future work and to the broader research community.

The need to explicitly control and disentangle formal inference from plausibility led us to focus mainly on a synthetically generated task. While our study on language modeling and OOD generalization explores the impact of steering beyond our setup, we believe further work is still required to fully understand and improve the steering methodology, assessing how steering for content effect can impact real-world applications in critical domains (e.g., science, healthcare).

\begin{figure*}[t]
\centering
\includegraphics[width=0.33\textwidth]{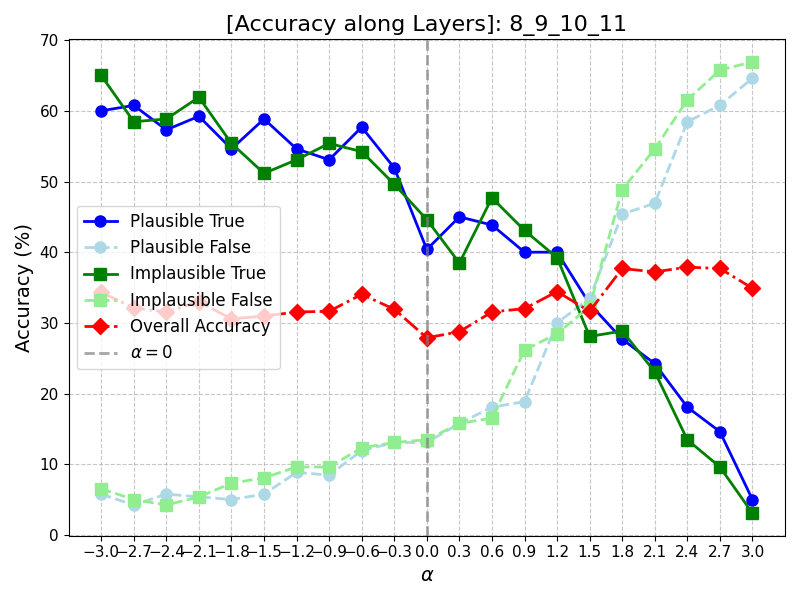}
\includegraphics[width=0.33\textwidth]{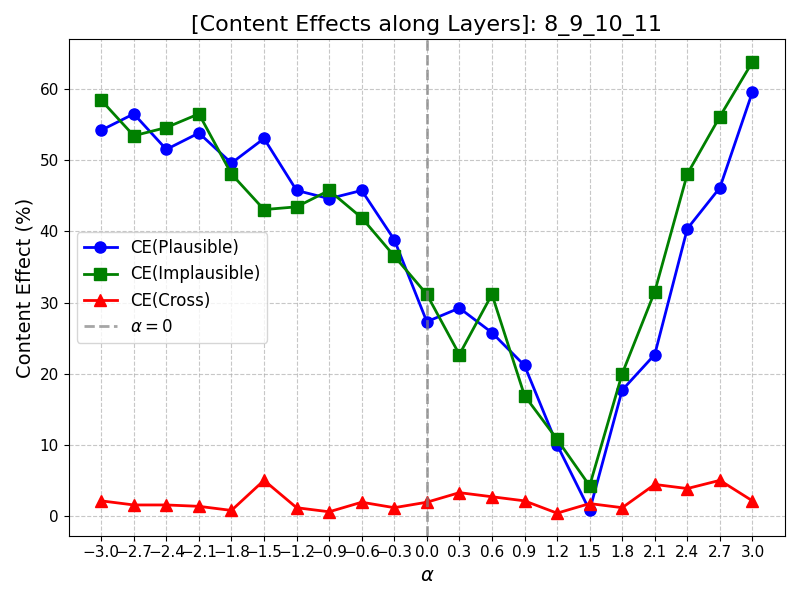}
\includegraphics[width=0.33\textwidth]{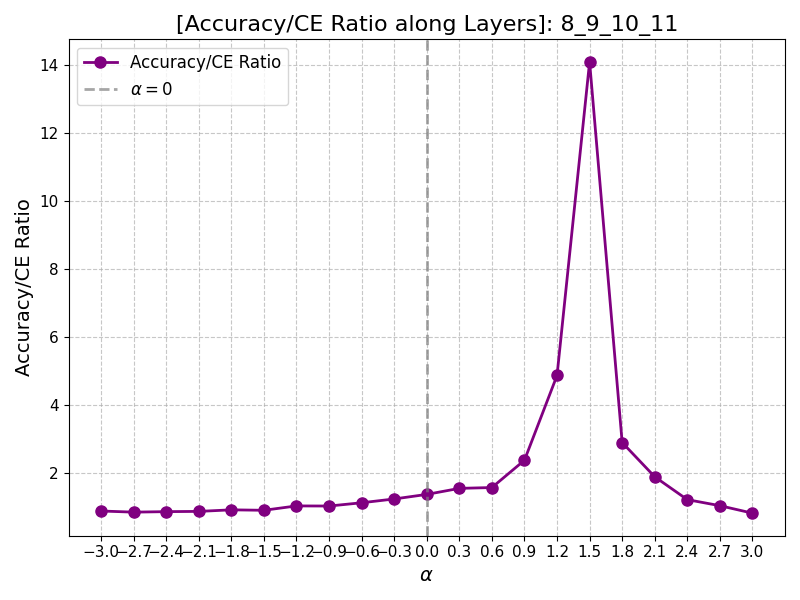}
\caption{Robustness of steering to prompt variations on Llama 1b (i.e., ACC, CE, and ACC/CE). The results reveal that, despite some noise deriving from perturbations applied at test time, the overall effectiveness of steering remains unaltered.}
\label{fig:robustness_llama}
\end{figure*}

\end{document}